\documentclass[letterpaper]{article} 
\usepackage{aaai24}  
\usepackage{times}  
\usepackage{helvet}  
\usepackage{courier}  
\usepackage[hyphens]{url}  
\usepackage{graphicx} 
\usepackage{natbib}  
\usepackage{caption} 
\usepackage{algorithm}
\usepackage{algorithmic}

\usepackage{microtype}
\usepackage{graphicx}
\usepackage{booktabs} 

\usepackage{subcaption}

\usepackage{amssymb}
\usepackage{multirow}
\usepackage{graphicx}
\usepackage{tabularx}
\usepackage[acronym]{glossaries}
\usepackage{makecell}
\usepackage{rotating}
\usepackage{threeparttable}
\usepackage{booktabs}
\usepackage{ragged2e}
\usepackage{array}
\usepackage{newfloat}
\usepackage{listings}
\DeclareCaptionStyle{ruled}{labelfont=normalfont,labelsep=colon,strut=off} 
\floatstyle{ruled}
\newfloat{listing}{tb}{lst}{}
\floatname{listing}{Listing}
\usepackage{amsmath}
\usepackage{amssymb}
\usepackage{mathtools}
\usepackage{amsthm}

\usepackage{amsmath,amsfonts,bm}

\def\eqref#1{equation~\ref{#1}}

\def\1{\bm{1}}

\DeclareMathAlphabet{\mathsfit}{\encodingdefault}{\sfdefault}{m}{sl}
\SetMathAlphabet{\mathsfit}{bold}{\encodingdefault}{\sfdefault}{bx}{n}

\usepackage[capitalize,noabbrev]{cleveref}

\theoremstyle{plain}

\theoremstyle{definition}

\theoremstyle{remark}

\newcommand{\bx}{\mathbf{x}}

\newcommand*\diff{\mathop{}\!\mathrm{d}}
\newcommand*\expm{\mathop{}\!\mathrm{exp}}
\newcommand*\logm{\mathop{}\!\mathrm{log}}

\title{Unified framework for diffusion generative models in SO(3): applications in computer vision and astrophysics}
 
 \author {
    Yesukhei Jagvaral\textsuperscript{\rm 1,2},
    Francois Lanusse\textsuperscript{\rm 3},
    Rachel Mandelbaum\textsuperscript{\rm 1,2}
}
\affiliations {
    \textsuperscript{\rm 1}  McWilliams Center for Cosmology, Department of Physics, Carnegie Mellon University, Pittsburgh, PA 15213 USA \\  
        \textsuperscript{\rm 2}  NSF AI Planning Institute for Data-Driven Discovery in Physics,  Carnegie Mellon University, Pittsburgh, PA 15213 USA \\  
    \textsuperscript{\rm3} 
 AIM, CEA, CNRS, Universite Paris-Saclay, Universite Paris Cite, 91191 Gif-sur-Yvette, France \\
    yjagvara@andrew.cmu.edu}

\usepackage{bibentry}

\begin{document}

\maketitle

\begin{abstract}
Diffusion-based generative models represent the current state-of-the-art for image generation. However, standard diffusion models are based on Euclidean geometry and do not translate directly to manifold-valued data. In this work, we develop extensions of both score-based generative models (SGMs) and Denoising Diffusion Probabilistic Models (DDPMs) to the Lie group of 3D rotations, SO(3). SO(3) is of particular interest in many disciplines such as robotics, biochemistry and astronomy/cosmology science. Contrary to more general Riemannian manifolds, SO(3) admits a tractable solution to heat diffusion, and allows us to implement efficient training of diffusion models. We apply both SO(3) DDPMs and SGMs to synthetic densities on SO(3) and demonstrate state-of-the-art results. Additionally, we demonstrate the practicality of our model on pose estimation tasks and in predicting correlated galaxy orientations for astrophysics/cosmology.
\end{abstract}

\section{Introduction }
Deep generative models (DGM) are trained to learn the underlying data distribution and then generate new samples that match the empirical data. There are several classes of deep generative models, including Generative Adversarial Networks \citep{gan}, Variational Auto Encoders \citep{vae} and Normalizing Flows \citep{n-flow}. Recently, a new class of DGMs based on Diffusion, such as Denoising Diffusion Probabilistic Models (DDPM) \citep{ho-ddpm} and Score Matching with Langevin Dynamics (SMLD) ,  a subset of general score-based generative models (SGMs), \citep{song2019}, have achieved state-of-the-art quality in generating images, molecules, audio and graphs\footnote{For a comprehensive list of articles on score-based generative modeling, see  \url{https://scorebasedgenerativemodeling.github.io/}} \citep{song2021}. Unlike GANs, training diffusion models is usually very stable and straightforward, they do not suffer as much from mode collapse issues, and they can generate images of similar quality.  
In parallel with the success of these diffusion models, \cite{song2021} demonstrated that both SGMs and DDPMs can mathematically be understood as variants of the same process. In both cases, the data distribution is progressively perturbed by a noise diffusion process defined by a specific Stochastic Differential Equation (SDE), which can then be time-reversed to generate realistic data samples from initial noise samples.
 
While the success of diffusion models has mainly been driven by data with Euclidean geometry (e.g., images), there is great interest in extending these methods to manifold-valued data, which are ubiquitous in many scientific disciplines. Examples include high-energy physics \citep{brehmer, manifold_qft}, astrophysics \citep{manifold_astro}, geoscience \citep{manifold_geo}, and biochemistry \citep{manifold_molecule}. Very recently, pioneering work has started to develop generic frameworks for defining SGMs on arbitrary compact Riemannian manifolds \citep{de_bortoli_riemannian_2022}, and non-compact Riemannian manifolds \citep{huang_riemannian_2022}.

In this work, instead of considering generic Riemannian manifolds, we are specifically concerned with the Special Orthogonal group in 3 dimensions, SO(3), which corresponds to the Lie group of 3D rotations. Modeling 3D orientations is of particularly high interest in many fields including for instance in robotics (estimating the pose of an object,  \citealt{robotics_pose}); and in biochemistry (finding the conformation angle of molecules that minimizes the binding energy, \citealt{molecular_conformations}).
 
Contrary to more generic Riemannian manifolds, SO(3) benefits from specific properties, including a tractable heat kernel and efficient geometric ODE/SDE solvers, that will allow us to define very efficient diffusion models specifically for this manifold.

The contributions of our paper are summarized as follows:
\begin{itemize}
    \item We reformulate Euclidean diffusion models on the SO(3) manifold, and demonstrate how the tractable heat kernel solution on SO(3) can be used to recover simple and efficient algorithms on this manifold.
    \item We provide concrete implementations of both Score-Based Generative Model and Denoising Diffusion Probabilistic Models specialized for SO(3).
    \item We reach a new state-of-the-art in sample quality on synthetic SO(3) distributions with our proposed SO(3) Score-Based Generative Model.
    \item We demonstrate the practicality and the expressive utility of our model in computer vision and astrophysics.
\end{itemize}

\section{Diffusion process on SO(3)}
 \label{sec:so3-diff}

   \newcolumntype{M}[1]{>{\centering\arraybackslash}m{#1}}
    \begin{figure*}
     
      \centering
        \begin{tabular}{M{165mm}  }
      \\
      \includegraphics[width=170mm]{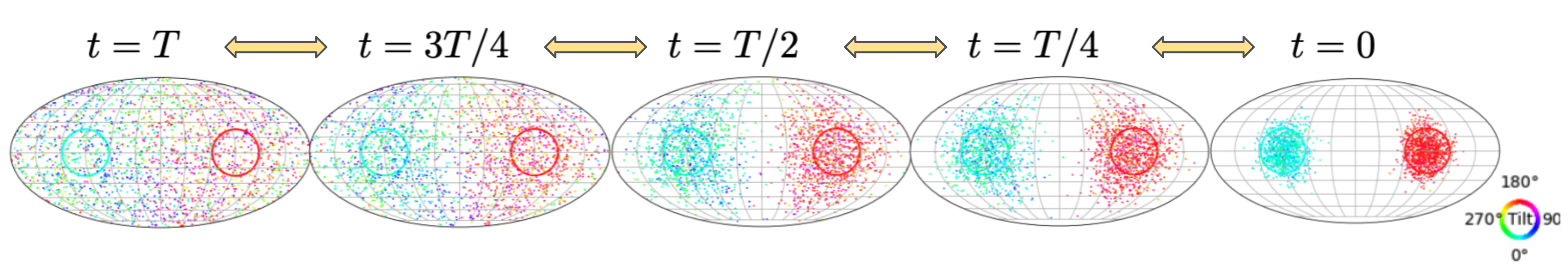} 
        \end{tabular}
             \caption{ \small Illustration of reversible diffusion of a mixture of two $\mathcal{IG}_\text{SO(3)}$ blobs on SO(3). Samples from a given base distribution (right most, denoted by circles) can be evolved under the probability flow ODE (Eq.~\ref{eqn:probability_flow_ODE}) towards a noisy distribution (left most), or vice-versa from the noisy distribution back to the target distribution. Each point represents a rotation matrix in SO(3) projected on the sphere according to its canonical axis, the color indicates the tilt around that axis (visualisation adopted from \citealt{Murphy21}). An animation of this figure is available* 
             }
   \label{fig:three graphs}
    \end{figure*}

In this work, we are exclusively considering the SO(3) manifold, corresponding to the Lie group of 3D rotation matrices. We will denote by $\expm: \mathfrak{so}(3) \rightarrow$SO(3) and $\logm : \text{SO(3)} \rightarrow \mathfrak{so}(3)$ the exponential and logarithmic maps that connect SO(3) to its tangent space and Lie algebra $\mathfrak{so}(3)$. 

Next, Similarly to Euclidean diffusion models \citep{song2021}, we begin by defining a Brownian noising process that will be used to perturb the data distribution. Let us assume a Stochastic Differential Equation of the following form:
\begin{equation}
    \diff \mathbf{x} = \mathbf{f}(\mathbf{x}, t)  \diff t + g(t) \diff \mathbf{w}, \label{eqn:forward_sde}
\end{equation}
where $\mathbf{w}$ is a Brownian process on SO(3),  $\mathbf{f}(\cdot \ , t): \text{SO(3)} \rightarrow T_{\mathbf{x}}$SO(3) is a drift term, and $g(\cdot): \mathbb{R} \rightarrow \mathbb{R}$ is a diffusion term, where $T_{\mathbf{x}}$SO(3) denotes the tangent space of SO(3). If we sample initial conditions for this SDE at $t=0$ from a given data distribution $\mathbf{x}(0) \sim p_\text{data}$, we will denote by $p_t$ the marginal distribution of $\mathbf{x}(t)$ at time $t > 0$. Thus $p_0=p_\text{data}$, and at final time $T$ at which we stop the diffusion process $p_T$ will typically tend to a known target distribution that will be easy to sample from.

Just like in the Euclidean case, as demonstrated in \citet{de_bortoli_riemannian_2022}, under mild regularity conditions \ref{eqn:forward_sde} admits a reverse diffusion process on compact Riemannian manifolds such as SO(3), defined by the following reverse-time SDE:
\begin{equation}
    \mathrm{d} \mathbf{x} = [\mathbf{f}(\mathbf{x}, t) - g(t)^2  \nabla \log p_t(\mathbf{x})] \mathrm{d} t + g(t) \mathrm{d} \bar{\mathbf{w}},\label{eqn:backward_sde}
\end{equation}
where $\bar{\mathbf{w}}$ is a reversed-time Brownian motion and the \textit{score function} $\nabla \log p_t(\mathbf{x}) \in T_{\mathbf{x}}$SO(3) is the derivative of the log marginal density of the forward process at time $t$. Corresponding to this reverse-time SDE, one can also define a probability flow ODE \citep{song2021}:
\begin{equation}
    \mathrm{d} \mathbf{x} = [\mathbf{f}(\mathbf{x}, t) - g(t)^2  \nabla \log p_t(\mathbf{x})] \mathrm{d} t .\label{eqn:probability_flow_ODE}
\end{equation}
This deterministic process is entirely defined once the score is known and maps $p_T$ to any intermediate marginal distributions $\{ p_t \}_{0 \leq t < T}$ of the forward process, including $p_0$. In particular, it can be seen as the equivalent of Neural ODE-based Continuous Normalizing Flows \citep[CNF,][]{chen_neural_2018} with an explicit parameterization in terms of the score function. We illustrate this process in Figure~\ref{fig:three graphs} with samples from two Gaussian-like blobs on SO(3) being transported reversibly through this ODE between $t=0$ and $t=T$.

While these equations are direct analog of the Euclidean SDEs and ODE described in \cite{song2021}, defining diffusion generative models on SO(3) will mainly differ on the two following points:
\begin{itemize}
    \item Defining the equivalent of the Gaussian heat kernel on SO(3): this is needed to easily sample from any intermediate $p_t$ without having to simulate an SDE.
    \item Solving SDEs and ODEs on the manifold: contrary to the Euclidean case, the diffusion process must remained confined to the SO(3) manifold, which requires specific solvers.
\end{itemize}
We address these two points below before moving on to defining our generative models on SO(3).

\subsection{The Isotropic Gaussian Distribution on SO(3)}
\label{sec:igso}

In general, the main disadvantage of working on Riemannian manifolds compared to Euclidean space is that they lack a closed form expression for the heat kernel, i.e., the solution of the diffusion process (which is a Gaussian in Euclidean space). For compact manifolds, the heat kernel is in general only available as an infinite series, which in the case of SO(3), takes the following form \citep{nikolayev70}:
\begin{equation}
 f_\epsilon(\omega) = \sum_{\ell=0}^{\infty} (2 \ell +1) \exp(- l (l+1) \epsilon^2) \frac{\sin((\ell + 1/2) \omega)}{\sin(\omega/2)} \label{eqn:heat_kernel}
\end{equation}
where $\omega = |\bm{\omega}| \in \left( -\pi, \pi \right]$ is the rotation angle of the axis-angle representation $\bm{\omega}$ of a given rotation matrix and $\epsilon$ is a concentration parameter.

While for $\epsilon > 1$ this series converges quickly ($\ell_{\max} = 5$ is sufficient to achieve sub-percent accuracy), the convergence gets slower as $\epsilon$ gets smaller, which makes it impractical to model concentrated distributions. Thankfully, this series has been thoroughly studied in the literature and \cite{Matthies80} shows that an excellent approximation of \ref{eqn:heat_kernel} can be achieved for $\epsilon < 1$ using the following closed-form expression:

\begin{multline} \label{eqn:approx_heat_kernel}
   f_\epsilon(\omega)  \simeq  \sqrt{\pi} \epsilon^{-3/2} e^{\frac{\epsilon}{4} - \frac{(\omega/2)^2}{\epsilon}} \\ \left(
    \frac{ \omega - e^{- \frac{\pi^2}{\epsilon}} 
     \left( (\omega - 2 \pi) e^{\pi \omega/\epsilon} +  (\omega + 2 \pi) e^{-\pi \omega/\epsilon}  \right) }{2 \sin(\omega/2)} \right)
\end{multline}

Therefore, in practical applications, one can switch between using a truncation of eq. \ref{eqn:heat_kernel} for $\epsilon \geq 1$ and the approximation eq. \ref{eqn:approx_heat_kernel} for $\epsilon < 1$.


Because of the property of being a solution of a diffusion process on SO(3), $f_\epsilon$ can be used to define the manifold equivalent of the Euclidean isotropic Gaussian distribution, which we will refer to as $\mathcal{IG}_\text{SO(3)}$, the Isotropic Gaussian on SO(3) \citep{leach22, ryu_equivariant_2022}, also known in the literature as the normal distribution on SO(3) \citep{nikolayev70, Matthies80}. For a given mean rotation $\bm{\mu} \in \text{SO(3)}$ and scale $\epsilon$, the probability density of a rotation $\mathbf{x} \in \text{SO(3)}$ under $\mathcal{IG}_\text{SO(3)}(\bm{\mu}, \epsilon)$ is given by:
\begin{equation}
    \mathcal{IG}_\text{SO(3)}( \mathbf{x} ; \bm{\mu}, \epsilon) = f_\epsilon( \arccos \left[ 2^{-1} (\text{tr}(\bm{\mu}^T \mathbf{x}) -1) \right] ) \;. \label{eqn:igso}
\end{equation}
Sampling from $\mathcal{IG}_\text{SO(3)}(\bm{\mu}, \epsilon)$ is achieved in practice by inverse transform sampling. The cumulative distribution function over angles needed to sample with respect to the uniform distribution on SO(3) can be evaluated numerically given integrating $\frac{1 - \cos(\omega)}{\pi} f_\epsilon(\omega)$ over $ \left( -\pi, \pi \right]$. To form a rotation matrix $\mathbf{x} \sim \mathcal{IG}_\text{SO(3)}( \cdot ; \bm{\mu}, \epsilon)$, one therefore first samples a rotation angle by inverse transform sampling given this CDF, then samples uniformly on $\mathcal{S}^2$ a rotation axis $\mathbf{v}$, yielding an axis-angle representation of a rotation matrix $\bm{\omega}=\omega \mathbf{v}$, which is then shifted by the mean of the distribution according to $\bm{x} = \bm{\mu} \expm(\bm{\omega})$.

An important property of $\mathcal{IG}_\text{SO(3)}(\mu, \epsilon)$, which sets it apart from other distributions on SO(3) (e.g. Bingham, Matrix Fisher, Wrapped Normal, more on this in Appendix \ref{dists_so3} ), is that it remains \textit{closed under convolution}, as a direct consequence of being the solution of a diffusion process. The convolution of two centered $\mathcal{IG}_\text{SO(3)}$ distributions of scale parameter $\epsilon_1$ and $\epsilon_2$ is an $\mathcal{IG}_\text{SO(3)}$ distribution of scale $\epsilon_1 + \epsilon_2$.
 
We will also note two interesting asymptotic behaviors. For large $\epsilon$, it tends to $\mathcal{U}_\text{SO(3)}$, the uniform distribution on SO(3), while for small $\epsilon$ the distribution $\mathcal{IG}_\text{SO(3)}(\mathbf{I}, \epsilon)$ can locally be approximated in the axis-angle representation of the tangent space by a normal distribution $\mathcal{N}(0, \sigma^2 \mathbf{I})$ in $\mathbb{R}^3$, with $\epsilon = \frac{\sigma^2}{2}$.

\subsection{Solving ODEs on SO(3)}

\begin{algorithm}[t]
\caption{Geometric ODE solver on SO(3) (Heun's method) for $\diff \mathbf{x} = \mathbf{f}(\mathbf{x}, t) \diff t$}
\label{alg:heun}
\begin{algorithmic}[1]
\REQUIRE Step size $h$, initial condition $\mathbf{x}_0$, time steps $\{t_n\}_{n=0}^{N}$, number of steps $N$
\FOR{$n \in \{0, \ldots, N-1 \} $}
\STATE $\mathbf{y}_1 = h \ \mathbf{f}( \mathbf{x}_n, \  t_n)$ 
\STATE $\mathbf{y}_2 = h \ \mathbf{f}( \expm(\frac{1}{2} \mathbf{y}_1) \mathbf{x}_n, \ t_n + \frac{1}{2}h)$
\STATE $\mathbf{x}_{n+1} = \expm(\mathbf{y}_2) \mathbf{x}_n$
\ENDFOR
\STATE{\textbf{return }} $\{ \mathbf{x}_{n} \}_{n=0}^{N}$
\end{algorithmic}
\end{algorithm}

Thanks to the existence of a tractable heat kernel on SO(3), the generative models we will define in the next section will not actually require us to solve the SDEs introduced at the beginning of this section, and we will only need to solve the probability flow ODE defined in 
 eq. \ref{eqn:probability_flow_ODE}.
 
Solving differential equations on manifolds can broadly be achieved using two distinct strategies, either projection methods using a Euclidean solver followed by a projection step onto the manifold, or intrinsic methods that rely on additional structure of the manifold to define an iteration that remains by construction on the manifold. In this work, we are concerned with SO(3), which is not only a compact Riemannian manifold, but also possesses a Lie group structure, which makes it amenable to efficient solvers. In particular, we will make use of the Runge-Kutta-Munthe-Kaas (RK-MK) class of algorithms and direct the interested reader to \cite{lie-methods} for a review of Lie group integrators. We adopt in practice the Lie group equivalent of Heun's method, which is one variant of RK-MK integrators, and we provide the details of this integrator in Algorithm~\ref{alg:heun}.

While we will not require it in practice, it is also possible to build SDE solvers on SO(3) with a similar strategy, and we point the interested reader for instance to the Geodesic Random Walk algorithm described in \cite{de_bortoli_riemannian_2022}.

\section{Diffusion Generative Models on SO(3)}

In this section, we present two different approaches to building generative models based on the forward and reverse diffusion processes presented in Section \ref{sec:so3-diff} resulting in SO(3) specific Score-Based Generative Models, and Denoising Diffusion Probabilistic Models.

   \newcolumntype{M}[1]{>{\centering\arraybackslash}m{#1}}
    \begin{figure*}
        \centering
        \begin{tabular}{M{30mm}M{30mm}M{30mm}M{30mm}M{30mm} }
      \\
        $t=T$ & $t=3T/4$ & $t=T/2$ & $t=T/4$ & $t=0$  \\
            \includegraphics[width=34mm]{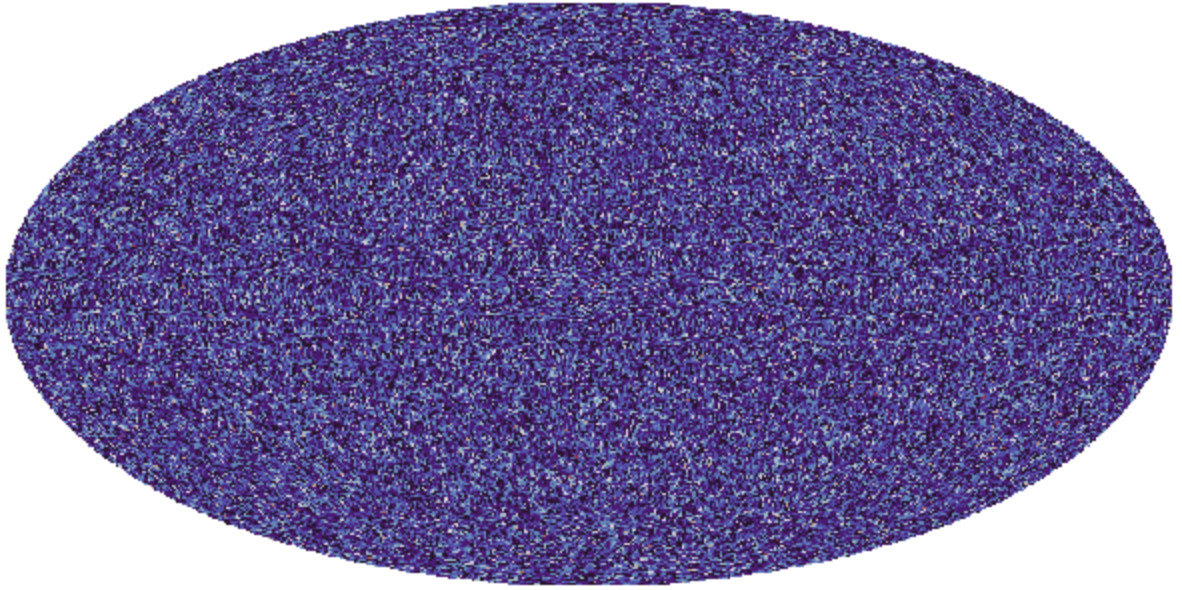} & \includegraphics[width=34mm]{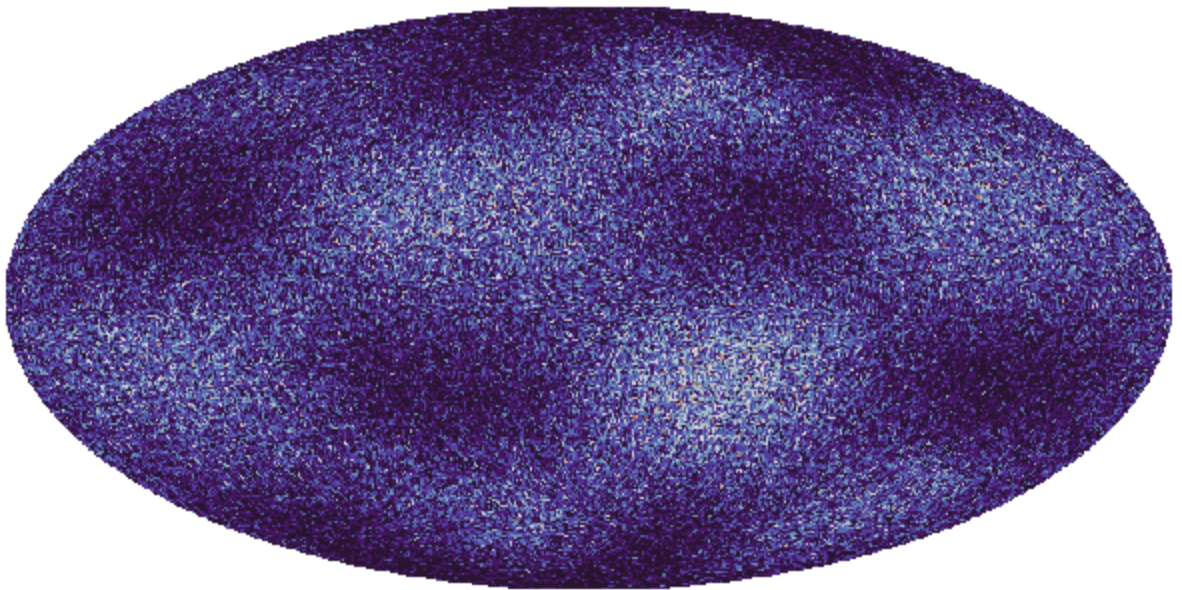} & \includegraphics[width=34mm]{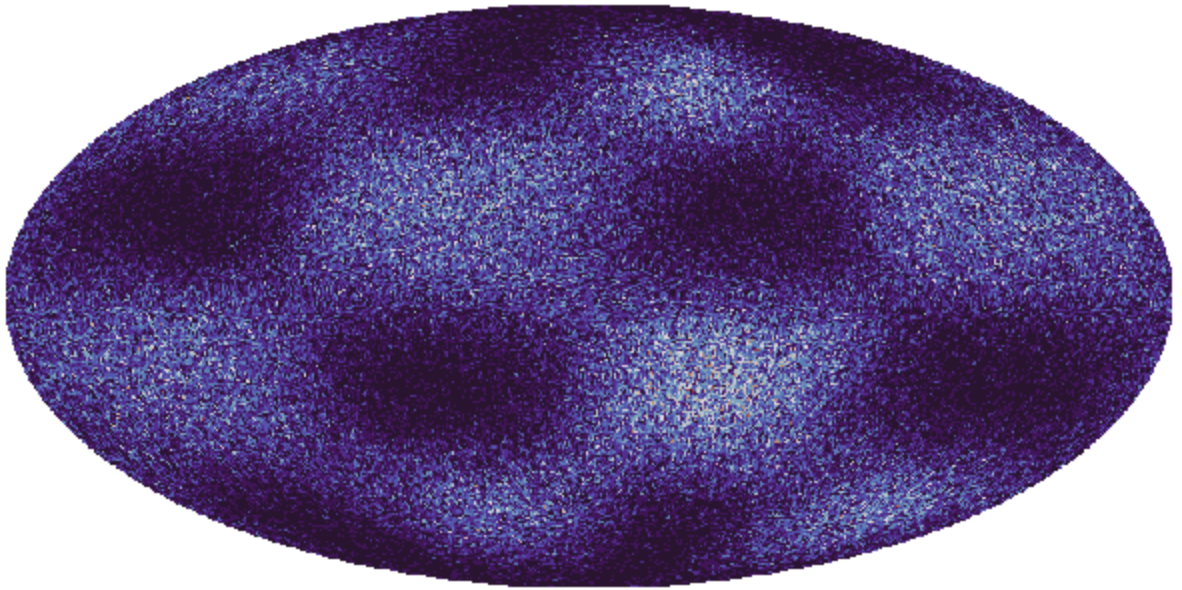} & \includegraphics[width=34mm]{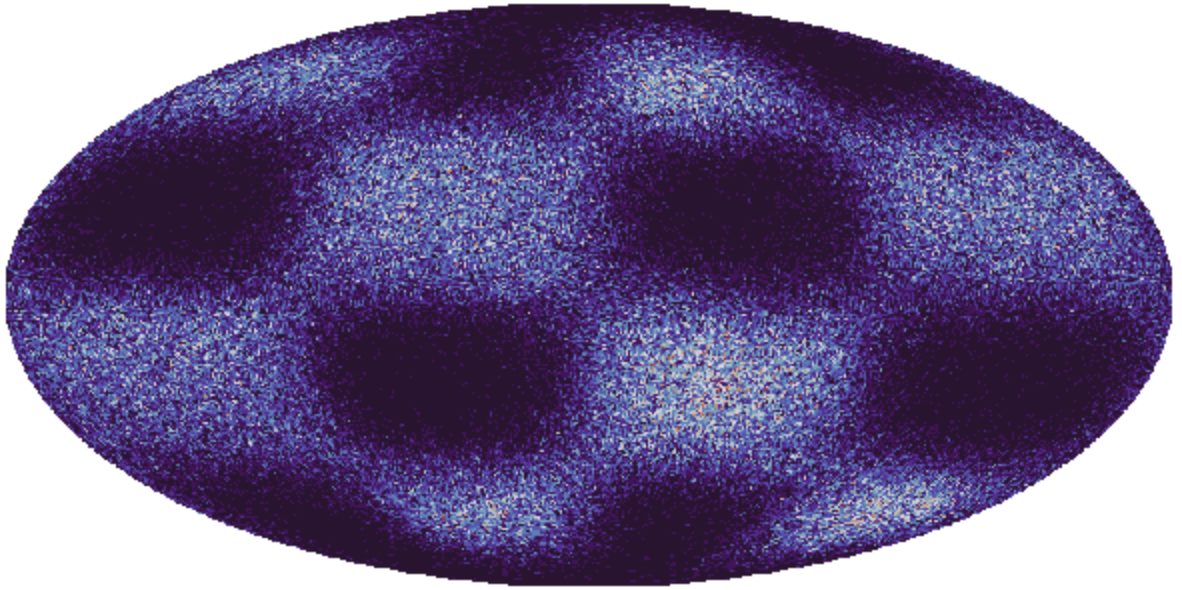} & \includegraphics[width=34mm]{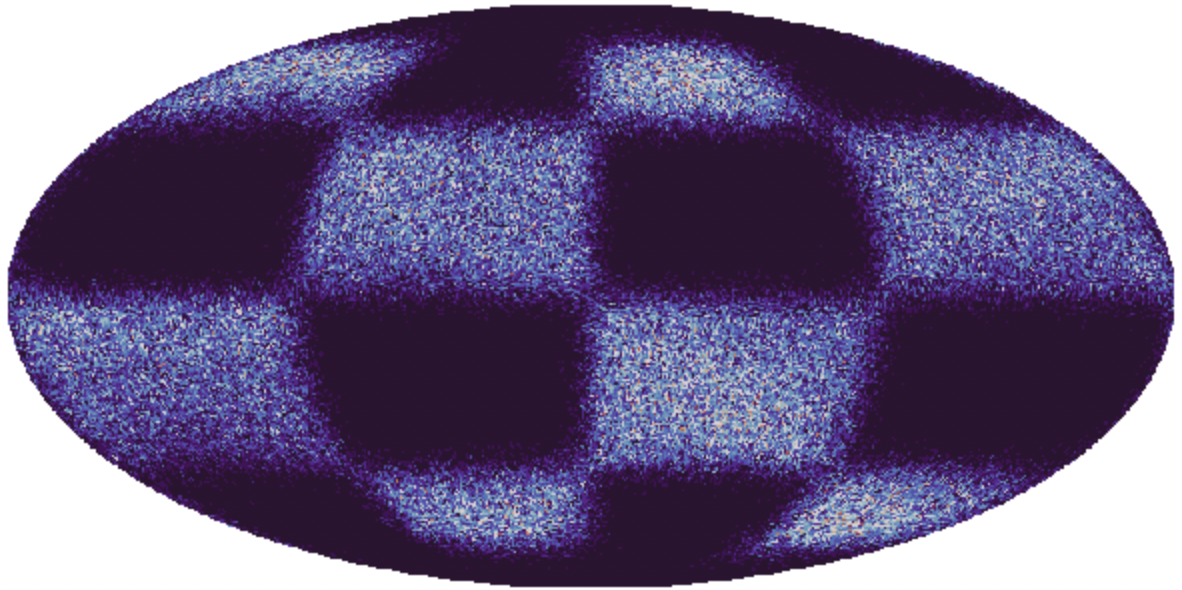}   \\
 
        \end{tabular}
        \caption{\small Sampling from a Diffusion Generative Model trained on a synthetic density on SO(3). Starting from $\mathcal{U}_{SO(3)}$, the uniform distribution on SO(3) at $t=T$ (left), the sampling procedure (either based on SGMs or DDPMs) denoises this distribution back to the target density at $t=0$ (right). For visualization this density plot shows the distribution of canonical axes of sampled rotations projected on the sphere; the tilt around that axis is discarded.}
        \label{tbl:toy_density}
    \end{figure*}

\subsection{Score-based Generative Model}

Although various choices for the particular form of the forward SDE in eq. \ref{eqn:forward_sde} are possible, for simplicity (and without loss of generality) we adopt in this section the so-called Variance-Exploding SDE, which is the canonical choice of the Euclidean Score-Matching Langevin Dynamics \citep{song2019}. More specifically, we define $\mathbf{f(\bm{x}, t)} = 0$ and $g(t)=\sqrt{\frac{\diff \epsilon(t)}{\diff t}}$ for a given choice of noise schedule $\epsilon(t)$, which reduces eq. \ref{eqn:forward_sde} to:
\begin{equation}
    \diff \mathbf{x} =\sqrt{\frac{\diff \epsilon(t)}{\diff t}} \diff \mathbf{w} \;. \label{eqn:ve_sde}
\end{equation}
For our fiducial model, and unless stated otherwise, we will further assume for simplicity the following noise schedule: $\epsilon(t)=t$.
The main drawback of this SDE in Euclidean geometry is that it will tend to a Gaussian with infinitely large variance. However, on SO(3) this SDE will tend to the uniform distribution $\mathcal{U}_\text{SO(3)}$, which is a natural choice for the prior distribution at large $T$.

Following from this choice of SDE, we can define a noise kernel $p_\epsilon(\tilde{\bm{x}} | \bm{x}) = \mathcal{IG}_\text{SO(3)}(\tilde{\bm{x}} ; \bm{x}, \epsilon)$ for $\bm{x}, \tilde{\bm{x}} \in$SO(3), such that the data distribution convolved by this noise kernel becomes
\begin{equation}
    p_\epsilon(\bm{x}) = \int_{SO(3)} p_\text{data}(\bm{x}^\prime) p_\epsilon(\bm{x} | \tilde{\bm{x}}) \diff \bm{x} \;,
\end{equation} 
and corresponds to $p_t$, the marginal distribution of the diffusion process at time $t$: $p_{\epsilon(t)} = p_t$.

This strategy directly extends Euclidean SGMs \citep{song2019, song2021} and relies on the time-reversed diffusion process described in eq. \ref{eqn:backward_sde}. Samples from the learned distribution $p_0$ can be sampled by first sampling $\mathbf{x}_T \sim \mathcal{U}_{SO(3)}$ and evolving these samples either through the reverse SDE  (eq. \ref{eqn:backward_sde}) or probability flow ODE  (eq. \ref{eqn:probability_flow_ODE}) back to $t=0$. This process is entirely defined as soon as the \textit{score function} of the marginal distribution at any intermediate time $t$, $\nabla \log p_{\epsilon(t)}$, is known. Therefore the first step is to establish a score-matching strategy on SO(3). 

Let us consider $\{ X_i \}_{i=0}^{3}$, an orthonormal basis of the tangent space $T_\mathbf{e}$SO(3). The directional derivative of the log density of the noise kernel $p_\epsilon(\bm{x} | \tilde{\bm{x}})$ can be computed as:
\begin{equation}
    \nabla_{X_i} \log p_\epsilon( \tilde{\mathbf{x}} | \mathbf{x}) =  \left. \frac{\mathrm{d}}{\mathrm{d} s} \log p_\epsilon( \tilde{\mathbf{x}} \expm(s X_i) | \mathbf{x}) \right|_{s=0} \;,
\end{equation}
which can be computed in practice by automatic differentiation given the explicit approximation formulae for the $\mathcal{IG}_\text{SO(3)}$ distribution introduced in Section \ref{sec:igso}.
To match this derivative, we introduce a neural score estimator $s_\theta(\mathbf{x}, \epsilon) : \text{SO(3)}\times\mathbb{R}^{+ \star} \rightarrow \mathbb{R}^3$, which can be trained directly under a conventional denoising score matching loss:
\begin{multline*}
    \mathcal{L}_{DSM} = \mathbb{E}_{p_\text{data}(\mathbf{x})} \mathbb{E}_{\epsilon \sim \mathcal{N}(0, \sigma_\epsilon^2)} \mathbb{E}_{p_{|\epsilon|}(\tilde{\mathbf{x}} | \mathbf{x} )} \\ \left[ |\epsilon| \ \parallel   s_\theta(\tilde{\mathbf{x}}, \epsilon) - \nabla_{X} \log p_{|\epsilon|}( \tilde{\mathbf{x}} | \mathbf{x}) \parallel_2^2 \right]
\end{multline*}
where we sample at training time random noise scales $\epsilon \sim \mathcal{N}(0, \sigma_\epsilon^2)$ similarly to \cite{song_improved_2020}. The minimum of this loss will be achieved for $s_\theta(\mathbf{x}, \epsilon) = \nabla \log p_{\epsilon}$. 

Once the score function is estimated from data using this score matching loss, sampling from the generative model can be achieved by using the reverse SDE formula, or using the ODE flow formula. In this work, we use the latter for its simplicity and speed, so that our specific fiducial sampling strategy becomes:
\begin{equation}
    \mathbf{x}_T \sim \mathcal{U}_\text{SO(3)}  \qquad ; \qquad \diff \mathbf{x}_t = -\frac{1}{2} \frac{\diff \epsilon(t)}{\diff t} s_\theta(\mathbf{x}_t, \epsilon(t)) \diff t
\end{equation}
which we solve down to $t=0$ with the geometric ODE solver described in Algorithm~\ref{alg:heun}. Compared to stochastic sampling strategies based on simulating the reverse SDE, this approach has several advantages. 1) It is much faster, and can benefit from adaptive ODE solvers bringing down the number of score evaluations needed, 2) the same ODE can be used to evaluate the log likelihood of the model by applying the probability flow formula of CNFs.

\subsection{Denoising Diffusion Probabilistic Model}
\label{sec:ddpm}

Similarly to the previous section, although several forms for the forward SDE in eq. \ref{eqn:forward_sde} are possible, to define a DDPM we adopt the canonical choice of \citet{ho-ddpm, sohl-dickstein} of a Variance-Preserving SDE defined as:
\begin{equation}
    \diff \mathbf{x} = -\frac{1}{2} \beta(t) \mathbf{x} \diff t + \sqrt{\beta(t)} \diff \mathbf{w}, \label{eqn:vp_sde}
\end{equation}
with $\beta(t)$ a function of time with values in $(0,1)$. This SDE will tend to a standard  $\mathcal{IG}_\text{SO(3)}$

\begin{algorithm}[tb]
\caption{Sampling from Denoising Diffusion Probabilistic Model on SO(3)}
\label{alg:algorithm_ddpm}
\begin{algorithmic}[1] 
\REQUIRE Trained neural networks $\mu_\theta(\bx, t), \epsilon_\theta(\bx, t)$, number of steps $N$, time steps $\{ t_i \}_{i=0}^{N}$ 
\STATE $\bx_{N} \sim \mathcal{IG}_\text{SO(3)}(\mathbf{I}, 1)$
\FOR {  $i=\{N, N-1, \ldots , 1\}$ }
        \STATE $\bx_{i-1} \sim p_\theta(\cdot ; \bx_{i}) = 
    \mathcal{IG}_\text{SO(3)}(\cdot ;  
     \mu_\theta(\bx_i, t_i) , \epsilon_\theta(\bx_i, t_i)) $   
 \ENDFOR 
 \STATE{\textbf{return }} $\{ \mathbf{x}_{n} \}_{n=0}^{N}$
 \end{algorithmic}
\end{algorithm}

As described in \citet{song2021}, when using a finite number of steps, the forward diffusion process defined by  eq. \ref{eqn:vp_sde} $\{\mathbf{x}_i\}_{i=0}^{N}$ (corresponding to times $\{0 \leq t_i \leq T\}_{i=0}^{n}$) can be interpreted as a Markov process:

\begin{equation}
    p(\bx_{0:N}) = p(\bx_0) p_{\beta_1}(\bx_1 | \bx_0)   \ldots p_{\beta_N}(\bx_N|\bx_{N-1})
\end{equation}

with the transition kernel $p_{\beta_{i+1}}(\bx_{i+1} | \bx_i) = \mathcal{IG}_\text{SO(3)}( \bx_{i+1} ; \sqrt{1 - \beta_{i+1}} \bx_i, \beta_{i+1})$.  The idea of DDPMs is to introduce a reverse Markov process defined in terms of variational transition kernels $p_\theta(\bm{x}_{i-1} | \bm{x}_{i})$:
\begin{equation}
    p_\theta(\mathbf{x}_{0:N}) = p_\theta(\bm{x}_N) p_\theta(\mathbf{x}_{N-1} | \mathbf{x}_{N}) \ldots p_\theta(\mathbf{x}_0| \mathbf{x}_1).
\end{equation}
While one could choose any distribution on SO(3) to parameterize this inverse transition kernel (e.g., Matrix Fisher, Bingham), we adopt for convenience an Isotropic Gaussian on SO(3) and use the following expression:
\begin{equation}
    p_\theta(\bm{x}_{i-1}|\bm{x}_i) = \mathcal{IG}_\text{SO(3)} ( \bm{x}_{i-1} ; \bm{x}_i \ \bm{\delta}_\theta(\bm{x}_i, t_i), \epsilon_\theta(\bm{x}_i, t_i) )
\end{equation}
where $\bm{\delta}_\theta : \text{SO(3)} \times \mathbb{R}^{+} \rightarrow \text{SO(3)}$ is a neural network predicting the residual rotation to apply to  $\bm{x}_i$ to obtain the mean of the reverse kernel and $\epsilon_\theta : \text{SO(3)} \times \mathbb{R}^{+} \rightarrow \mathbb{R}^+$ is a neural network predicting the variance of this reverse kernel. To parameterize the output of $\bm{\delta}_\theta$ we adopt the 6D continuous rotation representation of \citep{zhou2019continuity} and explore the impact of this choice in Section \ref{sec:repre}.

If the reverse Markov process can be successfully trained to match the forward process, it provides a direct sampling strategy to generate samples from $p_0$ by initializing the chain from $p_T$ and iteratively sampling from the reverse kernel $p_\theta(\bm{x}_{i-1}|\bm{x}_i)$. 
In DDPMs, the training strategy is to write down the Evidence Lower Bound (ELBO), given this variational approximation for the reverse Markov process, in order to train the individual transition kernels $p_{\theta} (\bx_{i-1}|\bx_i)$. To reduce the variance of this loss over a naive evaluation of the ELBO, \cite{sohl-dickstein} and \cite{ho-ddpm} propose to use a closed form expression of the reverse kernel $p(\bx_{{i-1}}| \bx_i, \bx_0)$ when conditioned on $\bx_0$. This makes it possible to rewrite the ELBO in terms of analytic KL divergences between Gaussian transitions kernels. 
However, contrary to the Gaussian case of Euclidean DDPMs, for $\mathcal{IG}_\text{SO(3)}$ we do not easily have access to a closed form expression of the reverse kernel $p(x_{t-1} | x_t, x_0)$ which is needed to derive the training loss used in \cite{ho-ddpm}. The same approach cannot be applied.

Instead, we consider the expression for the ELBO:
\begin{multline*}
    \mathbb{E}\left[ - \log p_\theta(\mathbf{x}_0) \right] \leq \\ \mathbb{E}_{p} \left[ - \log p(\mathbf{x}_N)  - \sum_{i\geq 1} \log \frac{p_\theta(\mathbf{x}_{i-1} | \mathbf{x}_{i} )}{p(\mathbf{x}_i | \mathbf{x}_{i-1})} \right] =: \mathcal{L_{\mathrm{ELBO}}}
\end{multline*}
which will be optimized by maximizing the log likelihood of individual transition kernels  $\log p_\theta(\mathbf{x}_{i-1} | \mathbf{x}_{i} )$ over samples $\mathbf{x}_{i-1}, \mathbf{x}_{i}$ obtained through simulating the forward Markov diffusion process over the training set.
Our strategy on SO(3), is therefore to train each transition kernel by maximum likelihood using the following loss function:
\begin{multline*}
    \mathcal{L}_{DDPM} := \\ \sum_{i \geq 0} \mathbb{E}_{p_\text{data}(\mathbf{x}_0)} \mathbb{E}_{p_\epsilon(\mathbf{x}_i | \mathbf{x}_0 )} \mathbb{E}_{p_{\epsilon_i}(\mathbf{x}_{i+1} | \mathbf{x}_i)} \left[ - \log p_\theta( \mathbf{x}_{i} | \mathbf{x}_{i+1}) \right]
\end{multline*}
where the log probability of the $\mathcal{IG}_\text{SO(3)}$ distribution used in our parameterised reverse kernel is defined in eq. \ref{eqn:igso}. While this loss can indeed be used to train a DDPM (as demonstrated in the next section), compared to the strategy of \cite{ho-ddpm}, we expect it to suffer from larger variance and is not explicitly parameterised in terms of the score function \citep{song2021}. Once trained, we can use the sampling strategy described in Algorithm~\ref{alg:algorithm_ddpm} to draw from the generative model.

\section{Experiments}

\begin{table*}
\centering

 \setlength\tabcolsep{5.5pt}
    \begin{tabular}{l c c c c c}
    \toprule
    Model & Checkerboard & 4-Gaussians & 3-Stripes\\
    \midrule
    
    SGM on SO(3) (ours) &  \textbf{0.50}{\scriptsize $\pm$ 0.01}  &  \textbf{0.50}{\scriptsize $\pm$ 0.01}  &  \textbf{0.51}{\scriptsize $\pm$ 0.01}\\
    DDPM on SO(3) (ours) &          0.52{\scriptsize $\pm$ 0.01}  &  0.53{\scriptsize $\pm$ 0.01}  &  0.52{\scriptsize $\pm$ 0.01}\\
    \hline
    RSGM \citep{de_bortoli_riemannian_2022} &           0.51{\scriptsize $\pm$ 0.01}  &  -- &  \textbf{0.51} {\scriptsize $\pm$ 0.01} \\
    Moser Flow \citep{rozen_moser_2021} &  0.56{\scriptsize $\pm$ 0.01}  &  0.60{\scriptsize $\pm$ 0.02}  &  0.53{\scriptsize $\pm$ 0.02} \\
    DDPM \citep{leach22} &           0.71{\scriptsize $\pm$ 0.04}  &  0.90{\scriptsize $\pm$ 0.05}  &  0.60{\scriptsize $\pm$ 0.03} \\
    Implicit-PDF \citep{Murphy21} &          0.59{\scriptsize $\pm$ 0.04}  &  0.81{\scriptsize $\pm$ 0.09}  &  0.63{\scriptsize $\pm$ 0.04} \\

    \bottomrule
    \end{tabular}
    \caption{\small Sample quality metric from the C2ST (lower is better). If the learned distribution is identical to the original one, the metric should be $\sim 0.5$; if it is significantly different, the metric tends towards $\sim 1$. The errors on the metric were obtain from the standard deviation of the metric over $k$-fold cross validation samples for a single training of the model. -- indicate a failure to evaluate the metric for a particular model.
    }\label{tab:c2st}
\end{table*}

\subsection{Test densities on SO(3)} 
\begin{figure*}[!ht]
     
      \centering
 
      \includegraphics[width=130mm]{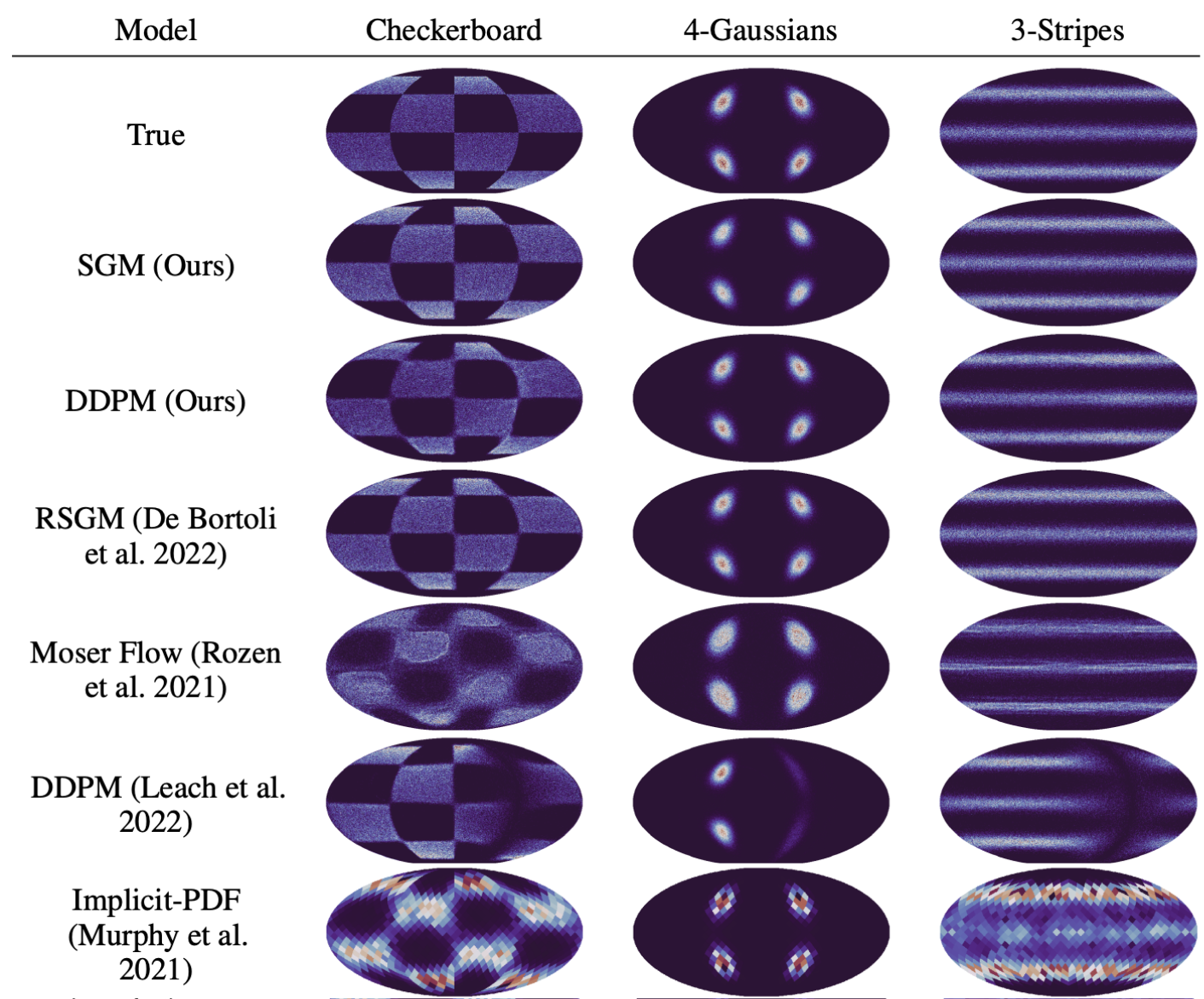} 
 
             \caption{ \small Density plot comparing samples from learned synthetic densities on SO(3). For visualization this density plot shows the distribution of canonical axes of sampled rotations projected on the sphere; the tilt around that axis is discarded. 
             }
   \label{fig:main_plot}
    \end{figure*}

We adopt three different toy distributions on SO(3): a checkerboard pattern, a multi-modal distribution of 4 concentrated Gaussians and a stripe pattern that can be viewed as circles on the sphere. We focus on evaluating the generative models in terms of the quality of their sample generation using the Classifier 2-Sample Tests (C2ST) metric \citep{c2st-lopez-paz2017revisiting,c2st-pmlr-v108-dalmasso20a,c2st-LueckmannBGGM21}. The C2ST metric has been used in particular in the context of simulation-based inference to quantify the quality of inferred distributions. Concisely, the C2ST method uses a neural network classifier to discriminate between true and the generated samples, yielding a value of $0.5$ if the two distributions are perfectly indistinguishable to the classifier, up to a value of $1$ if they are extremely different. In contrast to the usual Negative Log Likelihood (NLL), C2ST can be consistently computed for all generative models we compare bellow. \\
We present in  Figure \ref{fig:main_plot} and Table \ref{tab:c2st} the results of our comparisons on these test densities against the implicit-pdf method of \citet{Murphy21}, the DDPM implementation of \citet{leach22}, Moser flow of \citet{rozen_moser_2021}, and the Riemannian Score-Based Generative Model (RSGM) of \citet{de_bortoli_riemannian_2022} (trained under their $\ell_{t|0}$ score matching loss). We find that in all cases our SGM implementation on SO(3) yields the best C2ST metric, which is in line with the visual quality of distributions shown in Figure \ref{fig:main_plot}. Our DDPM implementation on SO(3) yields distributions that are comparatively less sharp, which we attribute to the larger variance of our training loss for that model. Compared to other models, our experiments illustrate a failure mode in the method of \citet{leach22} which we attribute to the fact that the usual DDPM loss function cannot be directly translated to SO(3) (as discussed in Section \ref{sec:ddpm}). We also note that the Implicit-PDF model, in comparison,  is extremely limited in resolution because of the memory cost of evaluating the pdf on a tiling of SO(3), and thus yields much lower scores. The best results after our method are achieved by the RSGM model \citep{de_bortoli_riemannian_2022}, which is expected due to its similarity with our work, but is slower to train in the specific case of SO(3). We find that the cost of simulating the forward SDE in the training phase leads to a factor x8 in computation time per batch on a given GPU.

\subsection{Pose estimation} 
\begin{figure} 
            \includegraphics[width=84mm]{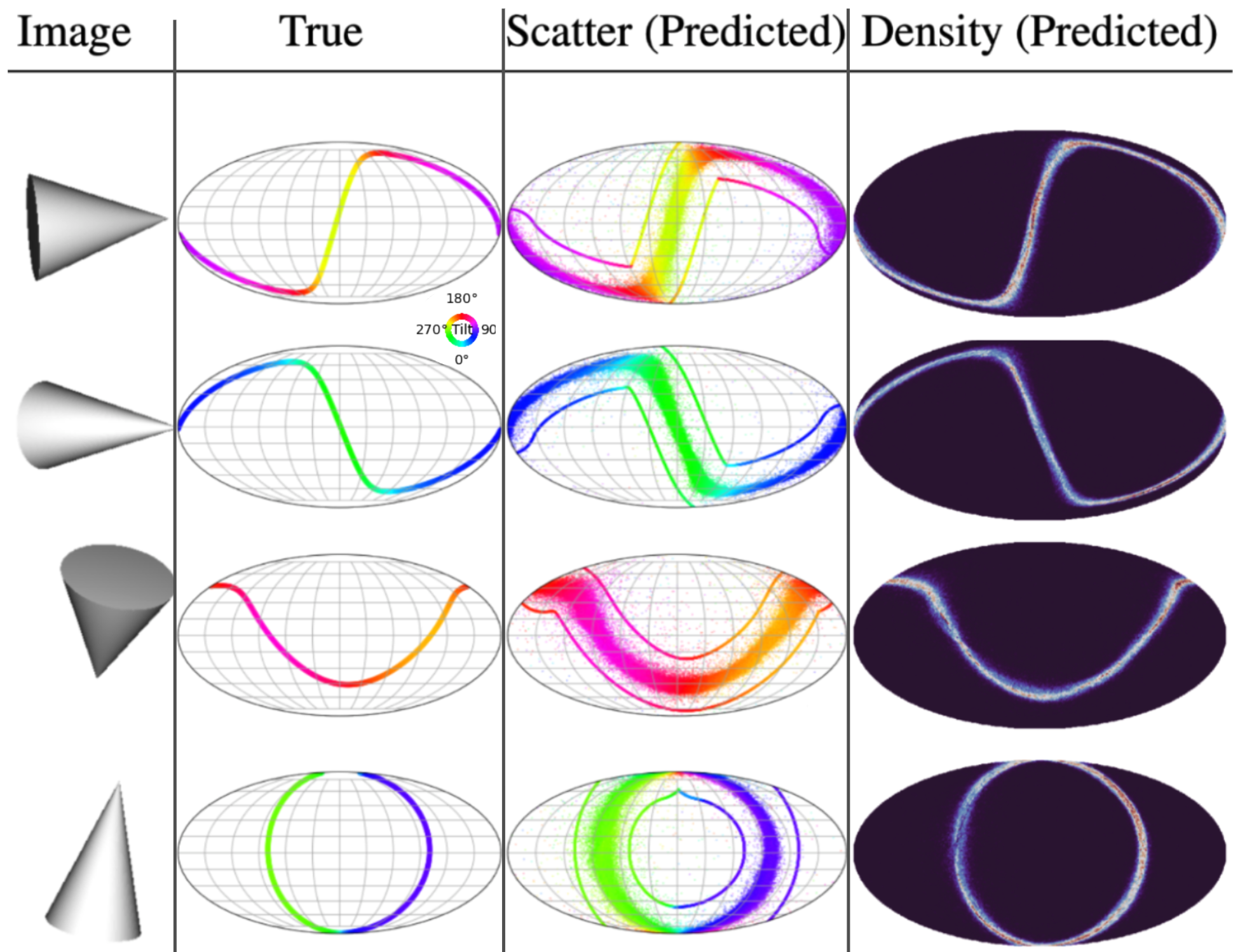}
        \caption{\small Predicted poses for an image of a solid with degenerate symmetry; here we only show results for a cone from multiple angles. The 1st column depicts the image of the symmetric solid and the 2nd column shows the underlying true pose of the object. In column 3, each point represents a rotation matrix in SO(3) projected on the sphere according to its canonical axis, with the
color indicating the tilt around that axis.
For visualization purposes, the density plot (column 4) shows the distribution of canonical axes of sampled rotations projected on the sphere; the tilt around that axis is discarded. 
        }
        \label{tbl:pose4}
    \end{figure}

To test practical applications of our model, 
following \citet{Murphy21} we used a vision description obtained from a pre-trained ResNet architecture with ImageNet weights consisting of 2048 dimensional vector to condition an SO(3) SGM. 
Conditioning the model on visual descriptors can be easily done by concatenating it to the input of the score network. 
Using images of symmetric solids from the SYMSOL dataset \citep{Murphy21} we show that we can correctly estimate poses of objects with degenerate symmetry, as shown in Figure \ref{tbl:pose4}. The model captures the underlying density well, as illustrated by column~4 where the probabilities are almost uniform along the density path. Additionally, uncertainty regions are expressed by the model as shown in column~3.  
Here we specifically chose symmetric objects in order to test the practical expressivity of our model. Pose estimation is more challenging for symmetric objects than for objects that lack symmetry, for which the process boils down to
learning and predicting a single point on SO(3) manifold. 
Compared to the Implicit-PDF results presented in \citet{Murphy21}, our approach necessitates far less memory, all the while reaching higher quality density estimates.

   \subsection{Conditional Galaxy Orientation Modeling}\label{3d_model}

   In this experiment, we are interested in emulating the behavior of galaxies in numerical simulations  of the Universe that follow the formation and evolution of galaxies  over cosmological times (from few hundred million years after the Big Bang  down to modern day). These simulations are computationally costly and even intractable at scales and resolutions demanded by future cosmological surveys. Therefore, emulation of galaxy properties using Machine  Learning methods could potentially significantly speed up analysis pipelines in cosmology. 
   The particular property we want to model is the 3D orientation of galaxies with respect to their surrounding environment. This important effect, called Intrinsic Alignment, relates to the fact that galaxies are not randomly oriented in the Universe, but tend to gain some preferential alignments through a variety of physical processes during their formation and evolution process.

   The problem can be formulated as a conditional density estimation problem over SO(3) given some summary information about their local environment, such as the local gravitational tidal field: given by the 3D tidal tensor $\bf{T}$  (more in Appendix \ref{appendix:TF}):
   
\begin{equation}
     p(\bf{x}_\text{galaxy}| \bf{T} ) 
\end{equation}
where $\bf{x}_\text{galaxy} \in \text{SO(3)}$  are the correlated galaxy orientations conditioned on the tidal field $\bf{T}$ around the galaxies. We want to model this density with a conditional score generative network $s_\theta(\bf{x}_\text{galaxy} | \bf{T})$ as such:
\begin{equation}
  \nabla_{\bf{x}_\text{galaxy}}  \mathrm{log} \, \big[ p(\bf{x}_\text{galaxy}|  \bf{T}) \big] \approx  s_\theta(\bf{x}_\text{galaxy} | \bf{T})
\end{equation}

\begin{figure}[!ht]\label{cent_gal}
 \centering
\includegraphics[height=7.0cm ]{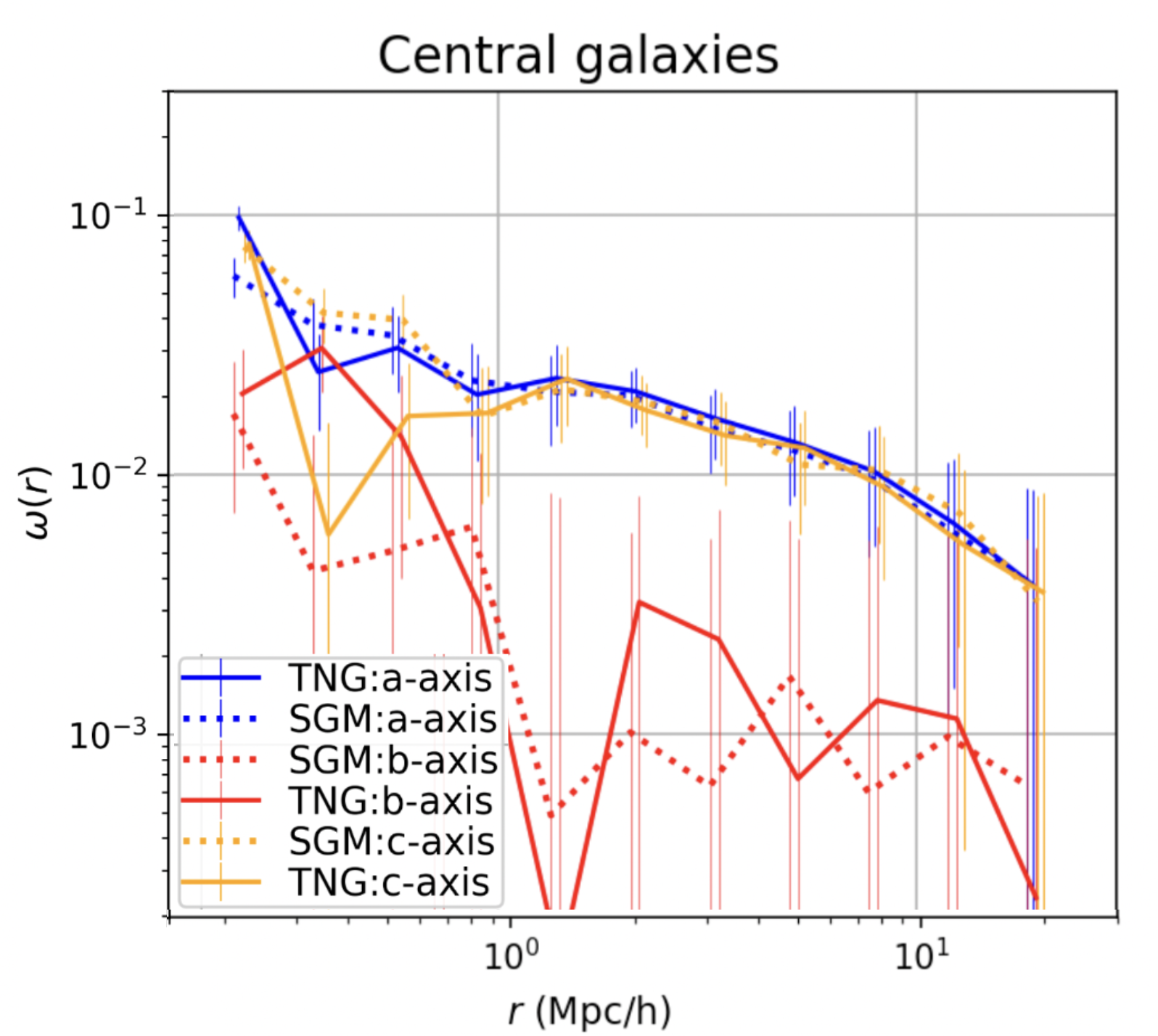}
\caption{\label{ED_halos}  \small
 The correlation function, $\omega(r)$, which captures the correlation between position and the axis direction,
of all galaxy   axes with galaxy positions: the solid lines show the measured values from the TNG simulation, while the dashed lines show the generated values from the SGM.
 The SGM curve was shifted by 5 per cent to the left for visual clarity. For the ellipsoid, we denote the major, intermediate, and minor axes as $a$, $b$, and $c$, respectively. Error bars were obtained using the jackknife method. }
 \end{figure}

\paragraph{Results  }\label{results} 
For this experiment we use  a state-of-the-art simulation, IllustrisTNG, (more in Appendix \ref{appendix:sim}), where we model galaxies as 3D ellipsoids (as it is conventional in astrophysics/cosmology), retrieve their 3D orientation and compute the summary information about their environments using $\bf{T}$. 
Throughout the section we  refer to the sample generated from the diffusion model as the \textit{SGM} sample, and the sample from  IllustrisTNG as the \textit{TNG} sample.  
 The inputs to the model are the  gravitational tidal field (obtained  from the 3D tidal tensor which carries some information about the alignment at large scales), and the outputs are the 3D orientations of  galaxies: the model generates the orientations of galaxies conditioned on the tidal field.

 In order to quantify how robustly we can recover the correct conditional density of orientation, we use a domain-specific quantitative measurement of correlation between galaxy orientations as a function of their respective distance -- $\omega(r)$  (more in Appendix \ref{appendix:ED}), and show very good agreement between the orientation of  galaxies with respect to the surrounding environment. This correlation function captures the correlation between  the large scale distribution of matter in the simulated volume (galaxy positions) and  orientations of the selected galaxy axes (modeling the galaxies as ellipsoids and selecting either the major, intermediate, or minor axis). Positive  $\omega(r)$ values indicate that the selected galaxy axis exhibits a coherent alignment towards the positions of  nearby galaxies.  The   $\omega(r)$ correlation functions for all three axes of the galaxies are presented in Fig.~\ref{ED_halos}.  
 In general, the qualitative trend of $\omega(r)$ as a function of 3D separation is captured by the SGM. For small  separation, there is a general deviation from the measured values, which may be explained by the highly complex hydrodynamical processes that might not have been captured by the neural network; or  the tidal field does not contain enough information. 
  Quantitatively, for the major and minor axes the generated samples agree well with the simulation.
   For the intermediate axes, the signal is very weak, though the SGM managed to captured the correlation with statistical consistency. 
    Overall, the SGM model can describe synthetic densities with high statistical correlations , and those with low statistical correlations, as shown in the case of galaxy alignments. 

\section{Conclusions and Limitations}
In this paper, we have presented a framework for score-based diffusion generative models on SO(3), as an extension of Euclidean SDE-based models \citep{song2021}. Because it is developed specifically for the SO(3) manifold, our work proposes a simpler and more efficient alternative to other recent (and general) Riemannian diffusion models while reaching state-of-the-art quality on synthetic distributions on SO(3). Additionally, we presented pose estimation results by conditioning the model on visual descriptors using the SYMSOL dataset. Our model correctly predicts the poses of objects with degenerate symmetry with low memory cost.
One of the most promising applications of this work is in robotics and computer vision, for the general task of pose-estimation, where our proposed model significantly outperforms current baselines \citep{Murphy21}. As for applications in the natural sciences, the generative model robustly captures low-signal statistical correlations of galaxy alignments in a state-of-the-art cosmological simulation.

\section*{Acknowledgements}
 This work was supported in part by a grant from the Simons Foundation (Simons Investigator in Astrophysics, Award ID 620789) and by the NSF AI Institute: Physics of the Future, NSF PHY- 2020295.

\bibliography{aaai24}

\begin{thebibliography}{53}
\providecommand{\natexlab}[1]{#1}

\bibitem[{Ben-Hamu et~al.(2022)Ben-Hamu, Cohen, Bose, Amos, Nickel, Grover,
  Chen, and Lipman}]{ben-hamu22a}
Ben-Hamu, H.; Cohen, S.; Bose, J.; Amos, B.; Nickel, M.; Grover, A.; Chen, R.
  T.~Q.; and Lipman, Y. 2022.
\newblock Matching Normalizing Flows and Probability Paths on Manifolds.
\newblock In Chaudhuri, K.; Jegelka, S.; Song, L.; Szepesvari, C.; Niu, G.; and
  Sabato, S., eds., \emph{Proceedings of the 39th International Conference on
  Machine Learning}, volume 162 of \emph{Proceedings of Machine Learning
  Research}, 1749--1763. PMLR.

\bibitem[{Brehmer and Cranmer(2020)}]{brehmer}
Brehmer, J.; and Cranmer, K. 2020.
\newblock Flows for Simultaneous Manifold Learning and Density Estimation.
\newblock In \emph{Proceedings of the 34th International Conference on Neural
  Information Processing Systems}, NIPS'20. Red Hook, NY, USA: Curran
  Associates Inc.
\newblock ISBN 9781713829546.

\bibitem[{Cao and Aziz(2020)}]{power-sph}
Cao, N.~D.; and Aziz, W. 2020.
\newblock The Power Spherical distribution.
\newblock \emph{ArXiv}, abs/2006.04437.

\bibitem[{Chen et~al.(2018)Chen, Rubanova, Bettencourt, and
  Duvenaud}]{chen_neural_2018}
Chen, R. T.~Q.; Rubanova, Y.; Bettencourt, J.; and Duvenaud, D.~K. 2018.
\newblock Neural {Ordinary} {Differential} {Equations}.
\newblock In \emph{Advances in {Neural} {Information} {Processing} {Systems}},
  volume~31. Curran Associates, Inc.

\bibitem[{Craven et~al.(2022)Craven, Croon, Cutting, and Houtz}]{manifold_qft}
Craven, S.; Croon, D.; Cutting, D.; and Houtz, R. 2022.
\newblock Machine learning a manifold.
\newblock \emph{Phys. Rev. D}, 105: 096030.

\bibitem[{Dalmasso et~al.(2020)Dalmasso, Lee, Izbicki, Pospisil, Kim, and
  Lin}]{c2st-pmlr-v108-dalmasso20a}
Dalmasso, N.; Lee, A.; Izbicki, R.; Pospisil, T.; Kim, I.; and Lin, C.-A. 2020.
\newblock Validation of Approximate Likelihood and Emulator Models for
  Computationally Intensive Simulations.
\newblock In Chiappa, S.; and Calandra, R., eds., \emph{Proceedings of the
  Twenty Third International Conference on Artificial Intelligence and
  Statistics}, volume 108 of \emph{Proceedings of Machine Learning Research},
  3349--3361. PMLR.

\bibitem[{De~Bortoli et~al.(2022)De~Bortoli, Mathieu, Hutchinson, Thornton,
  Teh, and Doucet}]{de_bortoli_riemannian_2022}
De~Bortoli, V.; Mathieu, E.; Hutchinson, M.; Thornton, J.; Teh, Y.~W.; and
  Doucet, A. 2022.
\newblock Riemannian {Score}-{Based} {Generative} {Modeling}.
\newblock ArXiv:2202.02763 [cs, math, stat].

\bibitem[{Falorsi et~al.(2019)Falorsi, de~Haan, Davidson, and
  Forré}]{falorsi_reparameterizing_2019}
Falorsi, L.; de~Haan, P.; Davidson, T.~R.; and Forré, P. 2019.
\newblock Reparameterizing {Distributions} on {Lie} {Groups}.
\newblock ArXiv:1903.02958 [cs, math, stat].

\bibitem[{Falorsi and Forré(2020)}]{falorsi_neural_2020}
Falorsi, L.; and Forré, P. 2020.
\newblock Neural {Ordinary} {Differential} {Equations} on {Manifolds}.
\newblock ArXiv:2006.06663 [cs, stat].

\bibitem[{Feiten, Lang, and Hirche(2013)}]{projected-gaussian}
Feiten, W.; Lang, M.; and Hirche, S. 2013.
\newblock Rigid motion estimation using mixtures of projected Gaussians.
\newblock In \emph{Proceedings of the 16th International Conference on
  Information Fusion}, 1465--1472.

\bibitem[{Gaddes, Hooper, and Bagnardi(2019)}]{manifold_geo}
Gaddes, M.~E.; Hooper, A.; and Bagnardi, M. 2019.
\newblock Using Machine Learning to Automatically Detect Volcanic Unrest in a
  Time Series of Interferograms.
\newblock \emph{Journal of Geophysical Research: Solid Earth}, 124(11):
  12304--12322.

\bibitem[{Gemici, Rezende, and Mohamed(2016)}]{gemici_normalizing_2016}
Gemici, M.~C.; Rezende, D.; and Mohamed, S. 2016.
\newblock Normalizing {Flows} on {Riemannian} {Manifolds}.
\newblock ArXiv:1611.02304 [cs, math, stat].

\bibitem[{Gilitschenski et~al.(2020)Gilitschenski, Sahoo, Schwarting, Amini,
  Karaman, and Rus}]{Igor2020}
Gilitschenski, I.; Sahoo, R.; Schwarting, W.; Amini, A.; Karaman, S.; and Rus,
  D. 2020.
\newblock Deep Orientation Uncertainty Learning based on a Bingham Loss.
\newblock In \emph{International Conference on Learning Representations}.

\bibitem[{Goodfellow et~al.(2014)Goodfellow, Pouget-Abadie, Mirza, Xu,
  Warde-Farley, Ozair, Courville, and Bengio}]{gan}
Goodfellow, I.; Pouget-Abadie, J.; Mirza, M.; Xu, B.; Warde-Farley, D.; Ozair,
  S.; Courville, A.; and Bengio, Y. 2014.
\newblock Generative Adversarial Nets.
\newblock In Ghahramani, Z.; Welling, M.; Cortes, C.; Lawrence, N.; and
  Weinberger, K., eds., \emph{Advances in Neural Information Processing
  Systems}, volume~27. Curran Associates, Inc.

\bibitem[{Grigoryan(2009)}]{grigoryan2009}
Grigoryan, A. 2009.
\newblock \emph{Heat Kernel and Analysis on Manifolds}.
\newblock AMS/IP studies in advanced mathematics. American Mathematical
  Society.
\newblock ISBN 9780821849354.

\bibitem[{Hartley et~al.(2013)Hartley, Trumpf, Dai, and Li}]{rotation-avg}
Hartley, R.~I.; Trumpf, J.; Dai, Y.; and Li, H. 2013.
\newblock Rotation Averaging.
\newblock \emph{International Journal of Computer Vision}, 103(3): 267--305.

\bibitem[{Hemmati et~al.(2019)Hemmati, Capak, Pourrahmani, Nayyeri, Stern,
  Mobasher, Darvish, Davidzon, Ilbert, Masters, and Shahidi}]{manifold_astro}
Hemmati, S.; Capak, P.; Pourrahmani, M.; Nayyeri, H.; Stern, D.; Mobasher, B.;
  Darvish, B.; Davidzon, I.; Ilbert, O.; Masters, D.; and Shahidi, A. 2019.
\newblock Bringing Manifold Learning and Dimensionality Reduction to {SED}
  Fitters.
\newblock \emph{The Astrophysical Journal}, 881(1): L14.

\bibitem[{Ho, Jain, and Abbeel(2020)}]{ho-ddpm}
Ho, J.; Jain, A.; and Abbeel, P. 2020.
\newblock Denoising Diffusion Probabilistic Models.
\newblock In Larochelle, H.; Ranzato, M.; Hadsell, R.; Balcan, M.; and Lin, H.,
  eds., \emph{Advances in Neural Information Processing Systems}, volume~33,
  6840--6851. Curran Associates, Inc.

\bibitem[{Hoque et~al.(2021)Hoque, Arafat, Xu, Maiti, and Wei}]{robotics_pose}
Hoque, S.; Arafat, M.~Y.; Xu, S.; Maiti, A.; and Wei, Y. 2021.
\newblock A Comprehensive Review on 3D Object Detection and 6D Pose Estimation
  With Deep Learning.
\newblock \emph{IEEE Access}, 9: 143746--143770.

\bibitem[{Huang et~al.(2022)Huang, Aghajohari, Bose, Panangaden, and
  Courville}]{huang_riemannian_2022}
Huang, C.-W.; Aghajohari, M.; Bose, A.~J.; Panangaden, P.; and Courville, A.
  2022.
\newblock Riemannian {Diffusion} {Models}.
\newblock ArXiv:2208.07949 [cs].

\bibitem[{Iserles et~al.(2000)Iserles, Munthe-Kaas, Nørsett, and
  Zanna}]{lie-methods}
Iserles, A.; Munthe-Kaas, H.~Z.; Nørsett, S.~P.; and Zanna, A. 2000.
\newblock Lie-group methods.
\newblock \emph{Acta Numerica}, 9: 215–365.

\bibitem[{Jagvaral et~al.(2022)Jagvaral, Lanusse, Singh, Mandelbaum,
  Ravanbakhsh, and Campbell}]{graphgan}
Jagvaral, Y.; Lanusse, F.; Singh, S.; Mandelbaum, R.; Ravanbakhsh, S.; and
  Campbell, D. 2022.
\newblock {Galaxies and haloes on graph neural networks: Deep generative
  modelling scalar and vector quantities for intrinsic alignment}.
\newblock \emph{Monthly Notices of the Royal Astronomical Society}, 516(2):
  2406--2419.

\bibitem[{{Kingma} and {Welling}(2013)}]{vae}
{Kingma}, D.~P.; and {Welling}, M. 2013.
\newblock {Auto-Encoding Variational Bayes}.
\newblock \emph{arXiv e-prints}, arXiv:1312.6114.

\bibitem[{Leach et~al.(2022)Leach, Schmon, Degiacomi, and Willcocks}]{leach22}
Leach, A.; Schmon, S.~M.; Degiacomi, M.~T.; and Willcocks, C.~G. 2022.
\newblock Denoising Diffusion Probabilistic Models on SO (3) for Rotational
  Alignment.
\newblock In \emph{ICLR 2022 Workshop on Geometrical and Topological
  Representation Learning}.

\bibitem[{{Lee} et~al.(2008){Lee}, {Springel}, {Pen}, and {Lemson}}]{ed-ee}
{Lee}, J.; {Springel}, V.; {Pen}, U.-L.; and {Lemson}, G. 2008.
\newblock {Quantifying the cosmic web - I. The large-scale halo
  ellipticity-ellipticity and ellipticity-direction correlations}.
\newblock \emph{Monthly Notices of the Royal Astronomical Society}, 389(3):
  1266--1274.

\bibitem[{Lopez-Paz and Oquab(2017)}]{c2st-lopez-paz2017revisiting}
Lopez-Paz, D.; and Oquab, M. 2017.
\newblock Revisiting Classifier Two-Sample Tests.
\newblock In \emph{International Conference on Learning Representations}.

\bibitem[{Lueckmann et~al.(2021)Lueckmann, Boelts, Greenberg, Gonçalves, and
  Macke}]{c2st-LueckmannBGGM21}
Lueckmann, J.-M.; Boelts, J.; Greenberg, D.~S.; Gonçalves, P.~J.; and Macke,
  J.~H. 2021.
\newblock Benchmarking Simulation-Based Inference.
\newblock In \emph{AISTATS}, 343--351.

\bibitem[{Mansimov et~al.(2019)Mansimov, Mahmood, Kang, and
  Cho}]{molecular_conformations}
Mansimov, E.; Mahmood, O.; Kang, S.; and Cho, K. 2019.
\newblock Molecular Geometry Prediction using a Deep Generative Graph Neural
  Network.
\newblock \emph{Scientific Reports}, 9.

\bibitem[{Marinacci et~al.(2018)}]{Marinacci2017illustristng}
Marinacci, F.; et~al. 2018.
\newblock {First results from the IllustrisTNG simulations: radio haloes and
  magnetic fields}.
\newblock \emph{Mon. Not. Roy. Astron. Soc.}, 480(4): 5113--5139.

\bibitem[{Mathieu and Nickel(2020)}]{mathieu_riemannian_2020}
Mathieu, E.; and Nickel, M. 2020.
\newblock Riemannian {Continuous} {Normalizing} {Flows}.
\newblock ArXiv:2006.10605 [cs, stat].

\bibitem[{Matthies, Muller, and Vinel(1988)}]{Matthies80}
Matthies, S.; Muller, J.; and Vinel, G.~W. 1988.
\newblock On the Normal Distribution in the Orientation Space.
\newblock \emph{Textures and Microstructures}, 10: 77--96.

\bibitem[{Mohlin, Bianchi, and Sullivan(2020)}]{mohlin_probabilistic_2020}
Mohlin, D.; Bianchi, G.; and Sullivan, J. 2020.
\newblock Probabilistic orientation estimation with matrix {Fisher}
  distributions.
\newblock ArXiv:2006.09740 [cs].

\bibitem[{Murphy et~al.(2021)Murphy, Esteves, Jampani, Ramalingam, and
  Makadia}]{Murphy21}
Murphy, K.~A.; Esteves, C.; Jampani, V.; Ramalingam, S.; and Makadia, A. 2021.
\newblock Implicit-PDF: Non-Parametric Representation of Probability
  Distributions on the Rotation Manifold.
\newblock In \emph{International Conference on Machine Learning}, 7882--7893.
  PMLR.

\bibitem[{{Naiman} et~al.(2018){Naiman}, {Pillepich}, {Springel},
  {Ramirez-Ruiz}, {Torrey}, {Vogelsberger}, {Pakmor}, {Nelson}, {Marinacci},
  {Hernquist}, {Weinberger}, and {Genel}}]{Naiman2018illustristng}
{Naiman}, J.~P.; {Pillepich}, A.; {Springel}, V.; {Ramirez-Ruiz}, E.; {Torrey},
  P.; {Vogelsberger}, M.; {Pakmor}, R.; {Nelson}, D.; {Marinacci}, F.;
  {Hernquist}, L.; {Weinberger}, R.; and {Genel}, S. 2018.
\newblock {First results from the IllustrisTNG simulations: a tale of two
  elements - chemical evolution of magnesium and europium}.
\newblock \emph{Monthly Notices of the Royal Astronomical Society}, 477(1):
  1206--1224.

\bibitem[{{Nelson} et~al.(2019){Nelson}, {Springel}, {Pillepich},
  {Rodriguez-Gomez}, {Torrey}, {Genel}, {Vogelsberger}, {Pakmor}, {Marinacci},
  {Weinberger}, {Kelley}, {Lovell}, {Diemer}, and {Hernquist}}]{tng-publicdata}
{Nelson}, D.; {Springel}, V.; {Pillepich}, A.; {Rodriguez-Gomez}, V.; {Torrey},
  P.; {Genel}, S.; {Vogelsberger}, M.; {Pakmor}, R.; {Marinacci}, F.;
  {Weinberger}, R.; {Kelley}, L.; {Lovell}, M.; {Diemer}, B.; and {Hernquist},
  L. 2019.
\newblock {The IllustrisTNG simulations: public data release}.
\newblock \emph{Computational Astrophysics and Cosmology}, 6(1): 2.

\bibitem[{Nelson et~al.(2018)}]{tng-bimodal}
Nelson, D.; et~al. 2018.
\newblock {First results from the IllustrisTNG simulations: the galaxy colour
  bimodality}.
\newblock \emph{Mon. Not. Roy. Astron. Soc.}, 475(1): 624--647.

\bibitem[{Nikolayev and Savyolov(1970)}]{nikolayev70}
Nikolayev, D.~I.; and Savyolov, T.~I. 1970.
\newblock Normal distribution on the rotation group SO (3).
\newblock \emph{Textures and Microstructures}, 29.

\bibitem[{Peretroukhin et~al.(2020)Peretroukhin, Giamou, Greene, Rosen, Kelly,
  and Roy}]{Valentin2020}
Peretroukhin, V.; Giamou, M.; Greene, W.~N.; Rosen, D.; Kelly, J.; and Roy, N.
  2020.
\newblock {A Smooth Representation of Belief over SO(3) for Deep Rotation
  Learning with Uncertainty}.
\newblock In \emph{Proceedings of Robotics: Science and Systems}. Corvalis,
  Oregon, USA.

\bibitem[{{Pillepich} et~al.(2018){Pillepich}, {Nelson}, {Hernquist},
  {Springel}, {Pakmor}, {Torrey}, {Weinberger}, {Genel}, {Naiman}, {Marinacci},
  and {Vogelsberger}}]{pillepich2018illustristng}
{Pillepich}, A.; {Nelson}, D.; {Hernquist}, L.; {Springel}, V.; {Pakmor}, R.;
  {Torrey}, P.; {Weinberger}, R.; {Genel}, S.; {Naiman}, J.~P.; {Marinacci},
  F.; and {Vogelsberger}, M. 2018.
\newblock {First results from the IllustrisTNG simulations: the stellar mass
  content of groups and clusters of galaxies}.
\newblock \emph{Monthly Notices of the Royal Astronomical Society}, 475(1):
  648--675.

\bibitem[{Rezende and Mohamed(2015)}]{n-flow}
Rezende, D.; and Mohamed, S. 2015.
\newblock Variational Inference with Normalizing Flows.
\newblock In Bach, F.; and Blei, D., eds., \emph{Proceedings of the 32nd
  International Conference on Machine Learning}, volume~37 of \emph{Proceedings
  of Machine Learning Research}, 1530--1538. Lille, France: PMLR.

\bibitem[{Rozen et~al.(2021)Rozen, Grover, Nickel, and
  Lipman}]{rozen_moser_2021}
Rozen, N.; Grover, A.; Nickel, M.; and Lipman, Y. 2021.
\newblock Moser {Flow}: {Divergence}-based {Generative} {Modeling} on
  {Manifolds}.
\newblock ArXiv:2108.08052 [cs, stat].

\bibitem[{Ryu et~al.(2022)Ryu, Lee, Lee, and Choi}]{ryu_equivariant_2022}
Ryu, H.; Lee, J.-H.; Lee, H.-i.; and Choi, J. 2022.
\newblock Equivariant {Descriptor} {Fields}: {SE}(3)-{Equivariant}
  {Energy}-{Based} {Models} for {End}-to-{End} {Visual} {Robotic}
  {Manipulation} {Learning}.
\newblock ArXiv:2206.08321 [cs].

\bibitem[{Sohl-Dickstein et~al.(2015)Sohl-Dickstein, Weiss, Maheswaranathan,
  and Ganguli}]{sohl-dickstein}
Sohl-Dickstein, J.; Weiss, E.; Maheswaranathan, N.; and Ganguli, S. 2015.
\newblock Deep Unsupervised Learning using Nonequilibrium Thermodynamics.
\newblock In Bach, F.; and Blei, D., eds., \emph{Proceedings of the 32nd
  International Conference on Machine Learning}, volume~37 of \emph{Proceedings
  of Machine Learning Research}, 2256--2265. Lille, France: PMLR.

\bibitem[{Song and Ermon(2019)}]{song2019}
Song, Y.; and Ermon, S. 2019.
\newblock Generative Modeling by Estimating Gradients of the Data Distribution.
\newblock In Wallach, H.; Larochelle, H.; Beygelzimer, A.; d\textquotesingle
  Alch\'{e}-Buc, F.; Fox, E.; and Garnett, R., eds., \emph{Advances in Neural
  Information Processing Systems}, volume~32. Curran Associates, Inc.

\bibitem[{Song and Ermon(2020)}]{song_improved_2020}
Song, Y.; and Ermon, S. 2020.
\newblock Improved {Techniques} for {Training} {Score}-{Based} {Generative}
  {Models}.
\newblock ArXiv:2006.09011 [cs, stat].

\bibitem[{Song et~al.(2021)Song, Sohl-Dickstein, Kingma, Kumar, Ermon, and
  Poole}]{song2021}
Song, Y.; Sohl-Dickstein, J.; Kingma, D.~P.; Kumar, A.; Ermon, S.; and Poole,
  B. 2021.
\newblock Score-Based Generative Modeling through Stochastic Differential
  Equations.
\newblock In \emph{International Conference on Learning Representations}.

\bibitem[{Springel et~al.(2018)}]{Springel2017illustristng}
Springel, V.; et~al. 2018.
\newblock {First results from the IllustrisTNG simulations: matter and galaxy
  clustering}.
\newblock \emph{Mon. Not. Roy. Astron. Soc.}, 475(1): 676--698.

\bibitem[{Srivatsan et~al.(2018{\natexlab{a}})Srivatsan, Xu, Zevallos, and
  Choset}]{Srivatsan-bingham}
Srivatsan, R.~A.; Xu, M.; Zevallos, N.; and Choset, H. 2018{\natexlab{a}}.
\newblock Probabilistic pose estimation using a Bingham distribution-based
  linear filter.
\newblock \emph{The International Journal of Robotics Research}, 37(13-14):
  1610--1631.

\bibitem[{Srivatsan et~al.(2018{\natexlab{b}})Srivatsan, Xu, Zevallos, and
  Choset}]{choset-bingham}
Srivatsan, R.~A.; Xu, M.; Zevallos, N.; and Choset, H. 2018{\natexlab{b}}.
\newblock Probabilistic pose estimation using a Bingham distribution-based
  linear filter.
\newblock \emph{The International Journal of Robotics Research}, 37(13-14):
  1610--1631.

\bibitem[{Thornton et~al.(2022)Thornton, Hutchinson, Mathieu, De~Bortoli, Teh,
  and Doucet}]{thornton_riemannian_2022}
Thornton, J.; Hutchinson, M.; Mathieu, E.; De~Bortoli, V.; Teh, Y.~W.; and
  Doucet, A. 2022.
\newblock Riemannian {Diffusion} {Schr}{\textbackslash}"odinger {Bridge}.
\newblock ArXiv:2207.03024 [cs, stat].

\bibitem[{von Mises(1918)}]{vmf}
von Mises, R. 1918.
\newblock Uber die 'Ganzzahligkeit' der Atomgewicht und verwandte Fragen.
\newblock \emph{Physikal. Z.}, 19: 490--500.

\bibitem[{Zelesko et~al.(2020)Zelesko, Moscovich, Kileel, and
  Singer}]{manifold_molecule}
Zelesko, N.; Moscovich, A.; Kileel, J.; and Singer, A. 2020.
\newblock Earthmover-Based Manifold Learning for Analyzing Molecular
  Conformation Spaces.
\newblock In \emph{2020 IEEE 17th International Symposium on Biomedical Imaging
  (ISBI)}, 1715--1719.

\bibitem[{Zhou et~al.(2019)Zhou, Barnes, Lu, Yang, and Li}]{zhou2019continuity}
Zhou, Y.; Barnes, C.; Lu, J.; Yang, J.; and Li, H. 2019.
\newblock On the continuity of rotation representations in neural networks.
\newblock In \emph{Proceedings of the IEEE/CVF Conference on Computer Vision
  and Pattern Recognition}, 5745--5753.

\end{thebibliography}

\newpage
\appendix

\section*{Appendix}

\section{Related work}
Most related to our work is \cite{song2021} which introduces the diffusion framework we use in this paper, and served as a point of reference throughout. We survey below related works that have developed methodologies to represent distributions on SO(3). 

\paragraph{Directional statistics} The classical approach for modeling distributions on SO(3) relies on (mixtures) of analytic distributions defined over the group of rotations. Common examples of using such distributions for modeling uncertainties over orientations include the Bingham distribution \citep{Valentin2020, Srivatsan-bingham, Igor2020} or the matrix Fisher distribution \citep{mohlin_probabilistic_2020}. The two main issues of these approaches are the lack of flexibility/expressivity of these analytic distributions, and the general difficulty of computing their normalization constant, which is typically required to train these models by maximum likelihood.

\paragraph{Normalizing Flows} A first class of methods proposes to use a conventional Euclidean Normalizing Flow in $\mathbb{R}^n$, which is then mapped to the target manifold using an invertible mapping \citep{gemici_normalizing_2016, falorsi_reparameterizing_2019}. This has some limitations however as the target manifold needs to be homeomorphic to $\mathbb{R}^{n}$ (which is the case for SO(3)), and this mapping can also present discontinuities. As an improvement over this approach, a second class of methods based on continuous normalizing flows \citep{chen_neural_2018} has emerged, defining directly flows on the manifold \citep{falorsi_neural_2020, mathieu_riemannian_2020}. These approaches remain relatively costly as training requires backpropagating through an ODE solver. \cite{rozen_moser_2021} proposes to sidestep that issue by training the CNF through penalizing the divergence of the neural network. And finally, in recent work \citep{ben-hamu22a} proposes to train a flow on manifolds by penalizing a Probability Path Divergence (PPD).

\paragraph{Diffusion models} In concurrent work, \cite{leach22} proposed an implementation of DDPMs on SO(3) by direct analogy with \cite{ho-ddpm}, based on the Isotropic Gaussian on SO(3) as a replacement for the Normal distribution in $\mathbb{R}^n$. However, as mentioned in the previous section, the loss function used in Euclidean DDPMs does not directly translate to SO(3), which leads to imperfect density estimation as we will illustrate in our experiments. 
Finally, \cite{de_bortoli_riemannian_2022, huang_riemannian_2022, thornton_riemannian_2022} introduce generic frameworks for diffusion models on Riemannian manifolds but only for Score-Based Generative Model (SGM). Their generic approach means they do not benefit from the knowledge of a solution to the heat equation in SO(3), which  we use extensively in our work to avoid the need to simulate SDEs and to efficiently generate samples from the forward diffusion process. In addition, we note that the method developed in \cite{huang_riemannian_2022} is not particularly efficient on the orthogonal group as it requires a projection operation, which involves a singular value decomposition.

\paragraph{Other approaches} \cite{Murphy21} develops a non-parametric representation of distributions on SO(3) by introducing a neural network to implicitly represent an unnormalized density on SO(3). Training this model by maximum likelihood requires computing the normalization constant of this implicit probability density function through brute-force evaluation on a tiling of SO(3), which is very costly in memory and limits the effective resolution of the learned densities.

Compared to the closest example in astrophysics to our work in \cite{graphgan}, our model can model the full 3D orientation, as opposed to only the major-axis. Additionally, our diffusion based architecture brings in all of the advantages of diffusion models compared to GANs, such as stability.

\section{Implementation and Training}
\label{appendix:implementation}

We designed our neural networks with a size of \{256,256,256,256,256\} neurons each with leaky ReLU activation. Additionally, the neural networks were conditioned on the noise scheduler and the noise scales were also learned parameters. 
We trained our models using the Adam optimizer with learning rate  of $10^{-4}$, exponential decay rates of $\beta_1=0.90$ and $\beta_2=0.95$, 400 000 iterations, and
a batch size of 1024. NVIDIA Tesla V100 GPU was used as the hardware, with JAX and DeepMind-Haiku Python libraries as the software. 

For the DDPM, we adopt in practice the Variance Preserving SDE of \citep{ho-ddpm} as we obtain better results empirically than with a Variance Exploding SDE.

\section{Representations of SO(3)} \label{representations_so3}
The special orthogonal group, SO(3), is the Lie group of all rotations about the origin in 3-dimensional space. There are several ways to represent the elements of the group SO(3), each with its advantages and disadvantages: 
\begin{itemize}
    \item Rotation Matrices  $\in \mathbb{ R}^{3\mathrm{x}3}$  with determinant equal to 1. This representation has 9 parameters and can be subject to some numerical stabilities, such as when computing the inverse or trigonometric functions.
    \item Euler angles (also called \textit{yaw, pitch, and roll} in robotics) 
    are three angles $\alpha, \beta, \psi$ that can describe an orientation with respect to a fixed coordinate system. This representation is subject to the infamous Gimbal lock, where one degree of freedom is lost when two axes of the gimbal become parallel. 
    \item Unit Quaternions defined as $\mathbf{\gamma} = a+b\mathbf{i}+c\mathbf{j}+d\mathbf{k}$, where $a,b,c,d$ are real number satisfying $\sqrt{a^2+b^2+c^2+d^2} =1 $ with $ \mathbf{i}+ \mathbf{j}+ \mathbf{k}$ denoting the vector (or imaginary) part of the unit quaternion. This representation has 4 parameters and has elegant operations (Hamilton product) without trigonometric functions \ref{quat_operations}. However, quaternions are antipodally symmetric which introduces some degeneracies.
    \item Axis - angle representation (normalized), Tangent space (unnormalized ) defined as  $\mathbf{\theta} = \theta \mathbf{e} = (\theta_1,\theta_2,\theta_3 ) = \theta ( e_1,e_2,e_3 ) $, where $\theta$ is the rotation angle and $\mathbf{e}$ is the rotation axis. However, this representation does not have well defined operations to combine rotations, and is furthermore discontinuous at $\theta=\pi$ \citep{zhou2019continuity}.

\end{itemize}
Therefore, in practice it is best to use some combinations of the aforementioned representations and convert back and forth among them. For a comprehensive review on SO(3) representations and metrics,  especially for  computer scientists, please refer to \citep{rotation-avg}.

\section{Impact of Rotation Representations on Neural Diffusion Model}\label{sec:repre}

As highlighted in \cite{zhou2019continuity}, a particular choice of rotation  representation can affect the training and accuracy of neural networks which either take rotations as an input or that output rotations. In particular, common representations such as axis-angle and quaternions are known to have discontinuities, which are needlessly difficult to capture for a neural network. In that work, they propose in particular to use 5 or 6 dimensional representations which have the particularity of being continuous and demonstrate their benefit in neural network training.

In our work, we make the choice of providing as inputs of the neural networks directly the 3x3 rotation matrix. Only the network involved in the DDPM needs to represent rotations as an output, and there we adopt the 6D representation following \cite{zhou2019continuity}, which can be seen as two 3D vectors, from which we can build a full orthogonal rotation matrix using cross-products.

\begin{figure}
     \centering
     \begin{subfigure}[b]{0.45\textwidth}
         \centering
         \includegraphics[width=\textwidth]{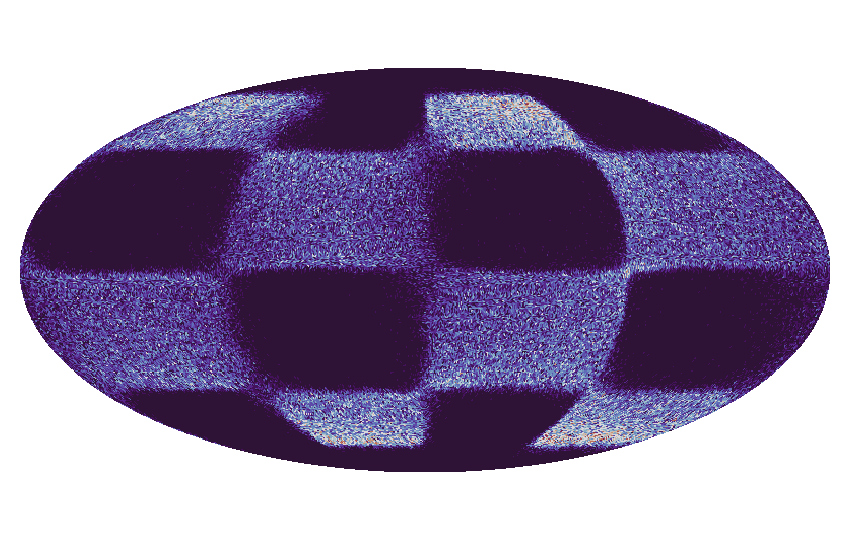}
         \caption{DDPM trained with axis-angle inputs and outputs.\\}
         \label{fig:axis-angle}
     \end{subfigure}
     \hfill
     \begin{subfigure}[b]{0.45\textwidth}
         \centering
         \includegraphics[width=\textwidth]{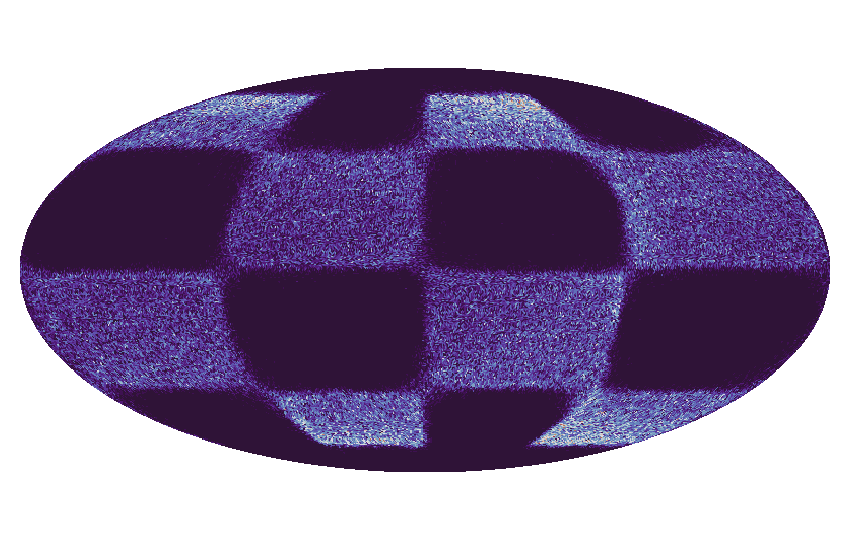}
         \caption{DDPM trained with 3x3 rotation matrices as inputs and using a continuous 6D output rotation representation.}
         \label{fig:6d}
     \end{subfigure}
        \caption{Comparison of distributions sampled from the DDPM under two different parameterizations of both input and output rotations. Once the models are fully trained as illustrated here, the impact is small, but on a partially trained network discontinuities would be visible in the case of the axis-angle representation, mostly due to the discontinuity in the input rotations.}
        \label{fig:representation}
\end{figure}

In comparing the impact of this choice against using only an axis-angle representation as inputs and outputs, we observe the following points:
\begin{itemize}
    \item The choice of the output parameterization (in the case of the DDPM) has no noticeable effect, which is expected as the model outputs residual rotations, which remain small and thus away from the discontinuity in the axis-angle representation.
    
    \item While the differences are small once the networks are fully trained, as illustrated on Figure \ref{fig:representation}, we notice for partially trained networks a discontinuity in the sampled distributions in the case of an input axis-angle representation. Therefore, we directly feed the 3x3 rotation matrix as an input to our networks.
\end{itemize}

\section{Quaternion operations for SO(3)} \label{quat_operations}

Quaternions form a group under multiplication, defined by the Hamilton product. Given quternions $\gamma_1$ and $\gamma_2$, the Hamilton product is defined by carrying out the $\gamma_1 \cdot \gamma_2 =  (a_1+b_1\mathbf{i}+c_1\mathbf{j}+d_1\mathbf{k}) \cdot (a_2+b_2\mathbf{i}+c_2\mathbf{j}+d_2\mathbf{k}) $ in a  distributive manner, keeping in mind the basis multiplication identities. This operation is physically equivalent to rotating by $\gamma_1$ and then by $\gamma_2$. The identity of the group is the quaternion $\gamma_0 = 1+ 0\mathbf{i}+0\mathbf{j}+0\mathbf{k}$ and the inverse of $\gamma*$ (also conjugate) is defined as $\gamma*=a-b\mathbf{i}-c\mathbf{j}-d\mathbf{k}$. 

 \paragraph{The reparametrization trick and variance preserving quaternions} In Euclidean space for  a Gaussian  random variable $\bx$ from 
 $\mathcal{N}(\mathbf{\mu};\sigma^2\mathbf{I})$, 
 the reparametrization trick is defined as $\bx= \mathbf{\mu} + \sigma^2 \cdot \delta$ 
 where $\delta \sim \mathcal{N}( 0 ; \textbf{I})$. We can define a analogous operation in the quaternion group, as such:
\begin{equation}
    \gamma = \theta \cdot  \epsilon^\delta
\end{equation}
Here, the quternion raised to some scalar power is defined as the $\gamma^a = \mathrm{exp}(\mathrm{ln}(\gamma) a ) $, in other words we take the quaternion to the tangent space from the manifold, perform the operation of multiplication and   bring it back into the manifold by using the exponential map.
and the variance preserving operation analogous to $\sqrt{\alpha} \cdot \bx + \sqrt{(1-\alpha)} \delta$ can be defined  as 
\begin{equation}
    \bx =  \bx^{\sqrt{\alpha} } \cdot \delta^{ \sqrt{(1-\alpha)} }
\end{equation}
For the variance exploding case, we can directly sample from the heat kernel without resorting to the quaternion operations. 

\subsection{Distributions on SO(3)} \label{dists_so3}

In literature there are numerous ways to represent distributions on the hypersphere. Most of them involve taking a standard distribution from the Euclidian space $\mathcal{R}^n$ and then constraining or projecting them on to the hypersphere $\mathcal{S}^n$. Some of the popular distributions are: 
\begin{itemize}
\item \textit{ Projected Gaussian(s)}  on the sphere   where standard Gaussian(s) on the tangent space of the hypersphere are projected via central projection, as done in \cite{projected-gaussian}; 

\item the \textit{ von Mises-Fisher} (vMF) distribution where an isotropic Gaussian on $R^n$ is restricted to the unit hypersphere \cite{vmf}; 
 
 \item The recently developed \textit{ Power Spherical} distribution \cite{power-sph}, which addresses some of the challenges of the vMF distribution, such as numerical stability  and scalability. 
 
 \item The antipodally symmetric \textit{ Bingham } distribution. The antipodal symmetry makes it a suitable distribution to represent quaternions, since quaternions double cover the space of rotations on SO(3) \cite{Igor2020, Valentin2020,choset-bingham, Srivatsan-bingham}. However, the Bingham distribution is notorious for its normalization constant that is very hard to compute. 
\end{itemize}
 Unfortunately, these distributions are not closed under convolution (i.e. composition of their random variables), thus writing down an closed form diffusion process akin to the Euclidean Gaussian case is intractable. 
One way to circumvent this problem  is to use class of functions that are closed under convolutions on the manifold. An obvious choice is the heat kernel which is the canonical solution to the diffusion equation and is closed under convolutions by definition \citep{grigoryan2009}. 

\section{Cosmological Simulation}\label{appendix:sim}
We have explored the efficacy of our model using the hydrodynamical TNG100-1 run at $z=0$ from the IllustrisTNG simulation suite \citep[for more information, please refer to][]{ tng-bimodal,pillepich2018illustristng, Springel2017illustristng, Naiman2018illustristng, Marinacci2017illustristng,tng-publicdata}. We employ a  stellar mass threshold of $ \log_{10}(M_*/M_\odot) \ge 9 $ for all galaxies, using the stellar mass from  their SUBFIND catalog, and select the central galaxies from each group for our analysis. The corresponding host dark matter halos were used to study halo alignments.

\section{Tidal field}\label{appendix:TF}
In order to predict galaxy and DM halo shapes and orientations, we use the tidal field defined as  the Hessian of the gravitational potential $\phi$:
\begin{equation}
    T_{ij}(\mathbf{r}) = \frac{\partial^2 \phi(\mathbf{r}) } {\partial r_i\partial r_j},
    \label{eq:tidal_field}
\end{equation}

To calculate the gravitational potential $\phi$, we start by computing the matter over-density field by constructing a particle mesh. First, the simulation box is divided into smaller 3D cubic cells. Within each cell $c$ centered at position  $\mathbf{r_c}$, we can count the total mass of the particles in that cell  and divided it by the average across all cells, 
and write the overdensity field as:
\begin{equation}
    \delta (\mathbf{r_c}) = \frac{\rho_c}{ \langle \rho_c \rangle} -1.
\end{equation}

The gravitational potential is related to the over-density field via the Poisson equation:
\begin{equation}
    \nabla^2\phi (\mathbf{r}) = 4\pi G \bar{\rho} \delta(\mathbf{r}),
    \label{eq:gravitational_potential}
\end{equation}
where $G$ is Newton's gravitational constant.
The solution of the Poisson equation in Fourier space is:
\begin{equation}
     \hat \phi(\mathbf{k}) = -4\pi G \bar{\rho} \frac{\hat\delta(\mathbf{k})}{k^2}  ,
\end{equation}

Plugging this result back into the Fourier transform of Eq.~\eqref{eq:tidal_field}, we obtain the tidal tensor 
\begin{equation}
     \hat T_{ij}(\mathbf{k}) = 4\pi G \bar{\rho} \frac{k_ik_j}{k^2}  \hat\delta(\mathbf{k}).
    \label{eq:tidal_field_k}
\end{equation}
In order to smooth the small scale coarseness of the tidal field (caused by the discrete resolution elements of the simulation), we introduce a Gaussian filter with smoothing scale $\gamma$:
\begin{equation}
     \hat{T}_{ij }(\mathbf{k}) = 4\pi G \bar{\rho} \frac{k_ik_j}{k^2} \hat\delta(\mathbf{k}) \;\; e^{-k^2 \gamma^2 /2}.
     \label{eq:tidal_fourier_gaussian}
\end{equation}

Finally, the Fourier-space tidal field from Eq.~\eqref{eq:tidal_fourier_gaussian} can be converted into real space using the inverse Fourier transform.
The tidal field was evaluated at the position of each galaxy, using a cloud-in-cell window kernel to interpolate between the centers of the grid points,
with various values of $\gamma$: 0.1 Mpc/h,  0.5 Mpc/h, 1 Mpc/h   on a mesh of size $1024^3$, with cell sizes given as $L_\text{box}/1024=0.073$ Mpc/h.

\section{Two-point correlation functions}\label{appendix:ED}
In order to quantifiably measure the correlations among the galaxy orientations we use the following standard estimator:
 The ellipticity-direction (ED) correlation function, which captures the position and the orientation correlation angles, defined as  \citep{ed-ee}: 
\begin{equation}
\omega(r) = \langle |\hat{e}({\bf x}) \cdot \hat{r}({\bf x})|^2 \rangle -\frac{1}{3} = \sum_{i \neq j} |\hat{e}_i \cdot \hat{r}_{ij}|^2  - \frac{1}{3}
\end{equation}
for a  galaxy at position $\bf x$ with ellipsoid axis direction $\hat{e}$ and the unit vector $\hat{r}$ denoting the direction of a density tracer at a distance $r$. This estimators captures whether galaxies tend to preferentially align themselves towards other galaxies, and the $\frac{1}{3}$ term was subtracted to remove the expected value.

\end{document}